%% file: main.tex
\documentclass[10pt,twocolumn,letterpaper]{article}

\usepackage[pagenumbers]{cvpr}      

\usepackage{graphicx}
\usepackage{amsmath}
\usepackage{amssymb}
\usepackage{booktabs}
\usepackage{array}
\usepackage{multicol}
\usepackage{multirow}
\usepackage{stackengine}
\usepackage{wrapfig}

\usepackage[pagebackref,breaklinks,colorlinks]{hyperref}

\usepackage[capitalize]{cleveref}
\crefname{section}{Sec.}{Secs.}
\Crefname{section}{Section}{Sections}
\Crefname{table}{Table}{Tables}
\crefname{table}{Tab.}{Tabs.}

\def\cvprPaperID{8965} 
\def\confName{CVPR}
\def\confYear{2023}

\begin{document}

\title{I$^2$-SDF: Intrinsic Indoor Scene Reconstruction and Editing\\ via Raytracing in Neural SDFs}


\author{
Jingsen Zhu\textsuperscript{1}
\and
Yuchi Huo\textsuperscript{2,1}
\and
Qi Ye\textsuperscript{3,6}
\and
Fujun Luan\textsuperscript{4}
\and
Jifan Li\textsuperscript{1}
\and
Dianbing Xi\textsuperscript{1}
\and
Lisha Wang\textsuperscript{1}
\and
Rui Tang\textsuperscript{5}
\and
Wei Hua\textsuperscript{2}
\and
Hujun Bao\textsuperscript{1}
\and
Rui Wang\textsuperscript{1}
\smallskip
\and
\textsuperscript{1}State Key Lab of CAD\&CG, Zhejiang University
\and
\textsuperscript{2}Zhejiang Lab
\and
\textsuperscript{3}Zhejiang University
\and
\textsuperscript{4}Adobe Research
\qquad
\textsuperscript{5}KooLab, Manycore
\qquad
\textsuperscript{6}Key Lab of CS\&AUS of Zhejiang Province
\\
\\
{\centering\tt\small\url{https://jingsenzhu.github.io/i2-sdf}}
}


\maketitle

\input{sections/abstract}

\input{sections/intro}

\input{sections/related}

\input{sections/overview}
\input{sections/method}
\input{sections/experiment}
\input{sections/conclusion}

{\small
\bibliographystyle{ieee_fullname}
\bibliography{egbib}
}

\newpage
\appendix
\input{sections/supp}

\end{document}

%% file: sections/abstract.tex
\begin{abstract}
   In this work, we present I$^2$-SDF, a new method for intrinsic indoor scene reconstruction and editing using differentiable Monte Carlo raytracing on neural signed distance fields (SDFs). Our holistic neural SDF-based framework jointly recovers the underlying shapes, incident radiance and materials from multi-view images. We introduce a novel bubble loss for fine-grained small objects and error-guided adaptive sampling scheme to largely improve the reconstruction quality on large-scale indoor scenes. Further, we propose to decompose the neural radiance field into spatially-varying material of the scene as a neural field through surface-based, differentiable Monte Carlo raytracing and emitter semantic segmentations, which enables physically based and photorealistic scene relighting and editing applications. Through a number of qualitative and quantitative experiments, we demonstrate the superior quality of our method on indoor scene reconstruction, novel view synthesis, and scene editing compared to state-of-the-art baselines. 
   Our project page is at \url{https://jingsenzhu.github.io/i2-sdf}.
\end{abstract}

%% file: sections/intro.tex
\section{Introduction}\label{sec:intro}

{
\setlength{\tabcolsep}{1pt}
\begin{figure}[t!]

    \centering
    \begin{tabular}{c|cc}
    \textsc{\small Rendering} & \multicolumn{2}{c}{\textsc{\small Scene Editing}}\\
    \includegraphics[height=2.0cm]{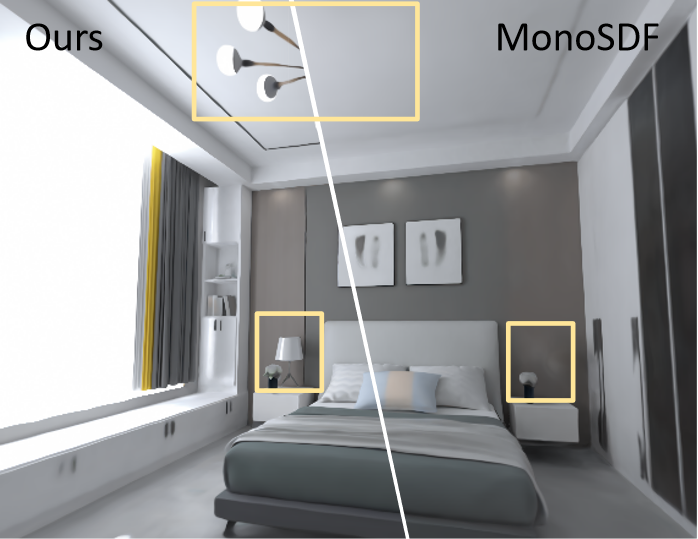} & \includegraphics[height=2.0cm]{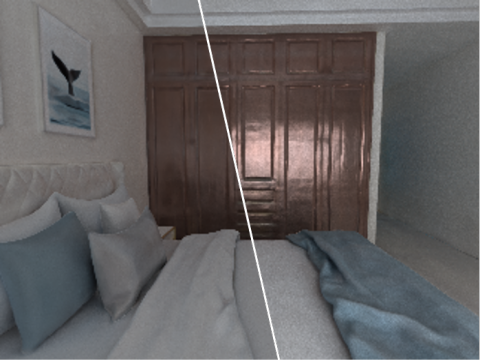} & 
    \stackinset{l}{1pt}{b}{1pt}{\includegraphics[height=0.7cm]{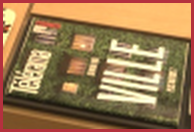}}{\includegraphics[height=2.0cm]{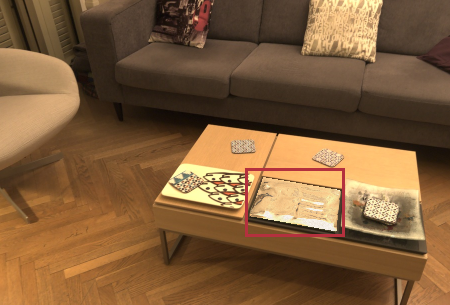}}
    \end{tabular}
    
    \caption{\textbf{I$^2$-SDF.} Left: State-of-the-art neural implicit surface representation method~\cite{Yu2022MonoSDF} fails in reconstructing small objects inside an indoor scene (\eg lamps and chandeliers), which is resolved by our bubbling method. Middle and Right: Our intrinsic decomposition and raytracing method enable photo-realistic scene editing and relighting applications.}
    \label{fig:teaser}
    \vspace{-2ex}
\end{figure}
}


Reconstructing 3D scenes from multi-view images is a fundamental task in computer graphics and vision. Neural Radiance Field (NeRF)~\cite{mildenhall2020nerf} and its follow-up research leverage multi-layer perceptions (MLPs) as implicit functions, taking as input the positional and directional coordinates, to approximate the underlying geometry and appearance of a 3D scene. Such methods have shown compelling and high-fidelity results in novel view synthesis. However, we argue that novel view synthesis itself is insufficient for scene editing such as inserting virtual objects, relighting and editing surface materials with global illumination.

On the other hand, inverse rendering or \emph{intrinsic decomposition}, which reconstructs and decomposes the scene into shape, shading and surface reflectance from single or multiple images, enables photorealistic scene editing possibilities. It is a long-term challenge especially for large-scale indoor scenes because they typically exhibit complex geometry and spatially-varying global illumination appearance. As intrinsic decomposition is an extremely ill-posed task, a physically-based shading model will crucially affect the decomposition quality. Existing neural rendering methods~\cite{boss2021nerd,zhang2021nerfactor,physg2021,munkberg2022extracting} rely on simple rendering algorithms (such as pre-filtered shading) for the decomposition and use a global lighting representation (e.g., spherical Gaussians). Although these methods have demonstrated the effectiveness on object-level inverse rendering, they are inapplicable to complex indoor scenes.
Moreover, indoor scene images are usually captured from the inside out and most lighting information has already presented inside the room. As a result, the reconstructed radiance field already provides sufficient lighting information without the need of active, external capture lighting setup.

To tackle the above challenges, we propose \emph{I$^2$-SDF}, a new method to decompose a 3D scene into its underlying shape, material, and incident radiance components using implicit neural representations. We design a robust two-stage training scheme that first reconstructs a neural SDF with radiance field, and then conducts raytracing in the SDF to decompose the radiance field into material and emission fields. As complex indoor scenes typically contain many fine-grained, thin or small structures with high-frequency details that are difficult for an implicit SDF function to fit, we propose a novel bubble loss and an error-guided adaptive sampling scheme that greatly improve the reconstruction quality on small objects in the scene. As a result, our approach achieves higher reconstruction quality in both geometry and novel view synthesis, outperforming previous state-of-the-art neural rendering methods in complex indoor scenes. Further, we present an efficient intrinsic decomposition method that decomposes the radiance field into spatially-varying material and emission fields using surface-based, differentiable Monte Carlo raytracing, enabling various scene editing applications. 

In summary, our contributions include:
\begin{itemize}
    \item We introduce {I$^2$-SDF}\footnote{I$^2$ meaning ``Intrinsics and Indoor''}, a holistic neural SDF-based framework for complex indoor scenes that jointly recovers the underlying shape, radiance, and material fields from multi-view images. 
    \item We propose a novel bubble loss and error-guided adaptive sampling strategy to effectively reconstruct fine-grained small objects inside the scene.
    \item We are the first that introduce Monte Carlo raytracing technique in scene-level neural SDF to enable photorealistic indoor scene relighting and editing.
    \item We provide a high-quality synthetic indoor scene multi-view dataset, with ground truth camera pose and geometry annotations.
\end{itemize}

%% file: sections/related.tex
\section{Related Work}

\paragraph{Neural implicit scene representations} (or \textit{neural fields})
have recently received extensive attention from the research community for representing 3D geometry and radiance information. Neural radiance field (NeRF)~\cite{mildenhall2020nerf} uses a single MLP to encode a scene as a continuous volumetric field of RGB radiance and density, giving promising results in novel view synthesis. Follow-up works accelerate reconstruction speed using voxels~\cite{SunSC22,yu_and_fridovichkeil2021plenoxels}, hashgrids~\cite{mueller2022instant} or deep image features~\cite{yu2020pixelnerf,mvsnerf}. Neural fields can also be applied to represent 3D geometric functions~\cite{Park_2019_CVPR,Oechsle2021ICCV,yariv2021volume,wang2021neus}. Despite their success in reconstructing small-scaled and textured objects, they have difficulties in handling shape-radiance ambiguity on texture-less surfaces. In this paper, we adopt one of the state-of-the-art implicit SDF methods, VolSDF~\cite{yariv2021volume}, as our neural implicit representation backbone and overcome the difficulties in indoor scene reconstruction task.


\paragraph{Neural 3D reconstruction for indoor scenes.}
Traditional multi-view stereo methods~\cite{schoenberger2016sfm, schoenberger2016mvs} can produce plausible geometry of textured surfaces, but struggle with texture-less regions such as white walls commonly seen in indoor scenes. Recently, learning-based MVS methods have been widely studied, which can be divided into two categories: depth-based methods and TSDF (truncated signed distance function) based methods. NeuralRecon~\cite{sun2021neucon} proposes a coarse-to-fine framework to regress input images to TSDF incrementally. NerfingMVS~\cite{wei2021nerfingmvs} leverages depth priors to guide the point sampling in NeRF to reduce shape-radiance ambiguity. NeRFusion~\cite{zhang2022nerfusion} combines the advantages of NeRF and TSDF-based fusion techniques to achieve reconstruction and rendering. Neural implicit SDF methods~\cite{wang2021neus,yariv2021volume}, which succeed in object-level 3D reconstruction, have also been applied to indoor scene reconstruction. To tackle with texture-less regions, additional priors are exploited to guide the network optimization, including semantic priors~\cite{guo2022manhattan}, normal priors~\cite{wang2022neuris,Yu2022MonoSDF} and depth priors~\cite{Yu2022MonoSDF}. The usage of priors results in improved reconstruction quality.


\paragraph{Inverse rendering} (also known as \emph{intrinsic decomposition}) is a long-term challenging and ill-posed problem in computer graphics and vision, which attempts to reconstruct and factorize the scene with geometry, material and lighting from single or multiple images. Monocular methods~\cite{li2018learning,garon2019fast,li2020inverse,wang2021learning,guo2021highlight,li2022physically,zhu2022learning} rely on strong priors from large-scale datasets. Recent methods exploit physically-based techniques such as differentiable rendering~\cite{li2020inverse,nimierdavid2021material} or raytracing~\cite{zhu2022learning} to achieve high-fidelity predictions, but cannot recover full 3D reconstructions that can be viewed from arbitrary viewpoints.
Multi-view methods holistically recover factorized 3D models for relighting and novel view synthesis from additional observations instead of strong priors. Neural implicit representations have been widely researched to estimate BRDF and lighting from image collections. NeRV~\cite{nerv2021} models light transport to support lighting effects such as shadows with high computational costs. Recent methods jointly estimate 3D geometry, BRDF and lighting from images.
Illumination is represented as spherical Gaussians (NeRD~\cite{boss2021nerd}, PhySG~\cite{physg2021}), low-resolution environment maps (NeRFactor~\cite{zhang2021nerfactor}), split-sum lighting model (Neural-PIL~\cite{boss2021neuralpil}, NVDIFFREC~\cite{munkberg2022extracting}), or Monte-Carlo estimator (NVDIFFREC-MC~\cite{hasselgren2022nvdiffrecmc}). However, these recent methods mainly focus on single object reconstruction and do not handle spatially-varying lighting conditions, which is not applicable to indoor scenes with complex geometry and lighting variations. In contrast, our method handles spatially-varying lighting and achieves indoor relighting with high fidelity.

%% file: sections/overview.tex
\section{Overview}

\begin{figure*}[ht!]
    \centering
    \vspace{-1ex}
    \includegraphics[width=\textwidth]{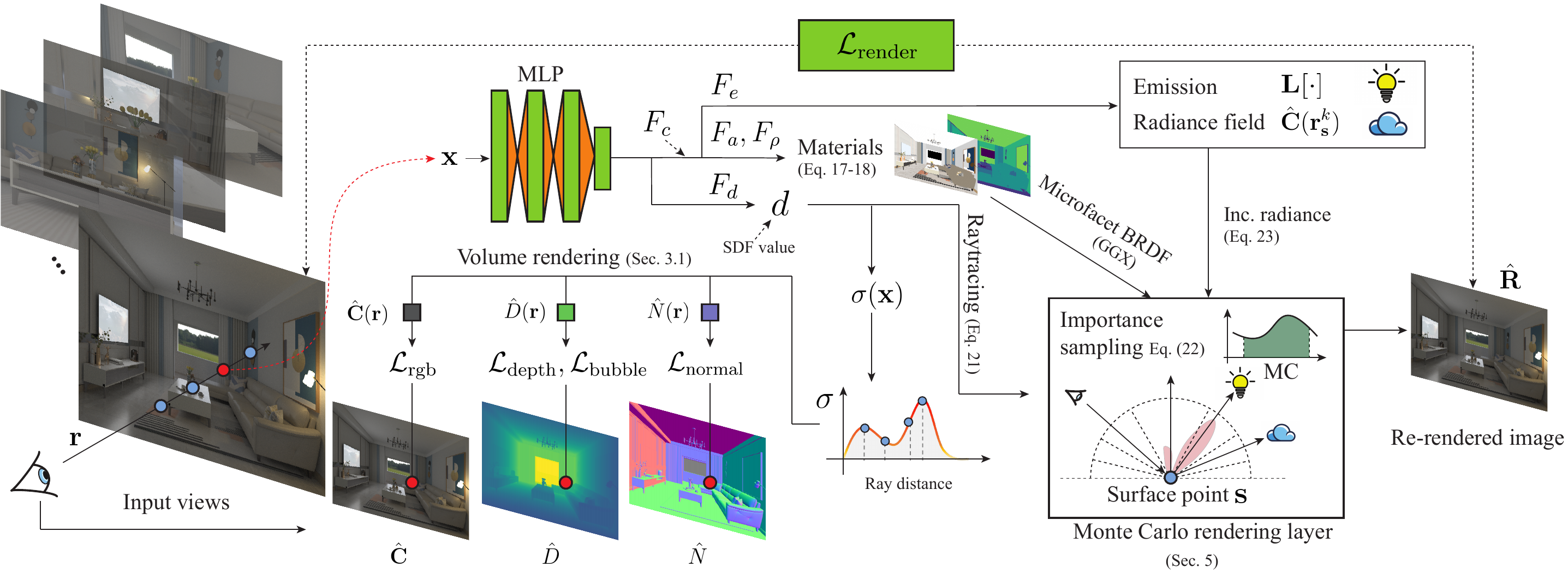}
    \vspace{-1.\baselineskip}
    \caption{\textbf{An overview of our pipeline.} Multi-view images are used to learn the underlying neural SDF field ($F_d$), radiance field ($F_c$), material fields ($F_a$ and $F_\rho$), and emission field ($F_e$ with $\mathbf{L}[\cdot]$), producing an intrinsic neural scene re-renderable for various applications.  }
    \label{fig:pipeline}
    \vspace{-1ex}
\end{figure*}


Our goal is to jointly decompose the underlying shape, incident radiance and material of the indoor scene according to multi-view input images with geometry priors such as depths and normals. 
We use implicit representations~\cite{mildenhall2020nerf,yariv2021volume} to model the scene geometry, radiance and material, parameterizing each factor as a single MLP. \cref{fig:pipeline} shows an overview of our pipeline, which consists of the neural SDF field ($F_d$), the neural radiance field ($F_c$), the neural material fields ($F_a$ and $F_\rho$) and emission field ($F_e$ with $\mathbf{L}[\cdot]$), and finally the Monte Carlo rendering layer which uses the decomposed factors to re-render the scene image.

To avoid the training ambiguities, we adopt a two-stage training scheme: We first train the geometry ($F_d$), radiance ($F_c$) and emitter semantic ($F_e$) fields, and then train the material ($F_a,F_\rho$) and emission ($\mathbf{L}[\cdot]$) fields. During the optimization of material and emission fields, $F_d,F_c,F_e$ are fixed and detached from the gradient descents.

In the following sections, we first review the concepts of neural SDF field and volume rendering in \cref{sec:sdf}. Next, we will respectively introduce the design details of decomposed components in \cref{sec:recon}, Monte Carlo rendering with raytracing in \cref{sec:render} and the training strategy in  \cref{sec:train}.

\subsection{Implicit Neural Surface Representation and Volume Rendering}\label{sec:sdf}

We represent the scene geometry as an implicit signed distance function (SDF). A signed distance function is a continuous function $d$ that maps a 3D point $\mathbf{x}$ to the closest distance of $\mathbf{x}$ to the surface:
\begin{equation}
    d: \mathbb{R}^3 \to \mathbb{R} \quad d(\mathbf{x}) = (-1)^{\mathbf{1}_\Omega(\mathbf{x})} \min_{\mathbf{y}\in\mathcal{M}}\|\mathbf{x}-\mathbf{y}\|,
\end{equation}
where $\Omega$ is the scene space and $\mathcal{M}=\partial\Omega$ is the scene surface. The sign of $d(\mathbf{x})$ indicates whether the point $\mathbf{x}$ is inside or outside the scene. In this work, we parameterize the SDF function as a single MLP $F_d$. Inspired by NeRF~\cite{mildenhall2020nerf}, we also parameterize the scene appearance as a view-dependent radiance field $F_c$:
\begin{align}
    (d(\mathbf{x}), \mathbf{z}(\mathbf{x})) = F_d(\mathbf{x}) \label{eq:sdf}, \\
    \mathbf{c}(\mathbf{x}) = F_c(\mathbf{z}(\mathbf{x}), \mathbf{v}),
\end{align}
where $\mathbf{z}(\mathbf{x})$ is a neural feature output by $F_d$ providing deep geometric cues to the radiance field $F_c$, and $\mathbf{v}$ is the view direction to model view-dependent visual effects such as specular reflections.

Following NeRF~\cite{mildenhall2020nerf}, we adopt differentiable volume rendering to learn scene implicit representation network from images. Specifically, to render a pixel, we cast a ray $\mathbf{r}$ from camera position $\mathbf{o}$ through the pixel along view direction $\mathbf{v}$. $M$ points $\mathbf{x}_{\mathbf{r}}^i=\mathbf{o}+t_{\mathbf{r}}^i \mathbf{v}$ are sampled along the ray and fed into $F_d$ and $F_c$ to predict their SDF value and radiance. In order to apply color accumulation, we transform SDF value $d_{\mathbf{r}}^i=d(\mathbf{x}_{\mathbf{r}}^i)$ to volume density $\sigma_{\mathbf{r}}^i=\sigma(\mathbf{x}_{\mathbf{r}}^i)$~\cite{yariv2021volume}:
\begin{equation}\label{eq:density}
    \sigma(\mathbf{x})= \begin{cases}\frac{1}{\beta}\left(1-\frac{1}{2} \exp \left(\frac{d(\mathbf{x})}{\beta}\right)\right) & \text { if } d(\mathbf{x})<0, \\ \frac{1}{2 \beta} \exp \left(-\frac{d(\mathbf{x})}{\beta}\right) & \text { if } d(\mathbf{x}) \geq 0,\end{cases}
\end{equation}
where $\beta$ is a learnable parameter to control the sparsity near the surface. The pixel color $\hat{\mathbf{C}}(\mathbf{r})$ for ray $\mathbf{r}$ is rendered via numerical integration~\cite{mildenhall2020nerf}:
\begin{equation}\label{eq:rgb}
    \hat{\mathbf{C}}(\mathbf{r})=\sum_{i=1}^M T_{\mathbf{r}}^i \alpha_{\mathbf{r}}^i \hat{\mathbf{c}}_{\mathbf{r}}^i,
\end{equation}
where $\delta_{\mathbf{r}}^i$ is the distance between adjacent sampled points $\mathbf{x}_{\mathbf{r}}^i$ and $\mathbf{x}_{\mathbf{r}}^{i+1}$, $\alpha_{\mathbf{r}}^i=1-\exp \left(-\sigma_{\mathbf{r}}^i \delta_{\mathbf{r}}^i\right)$ is the alpha value of each point, and $T_{\mathbf{r}}^i=\prod_{j=1}^{i-1}\left(1-\alpha_{\mathbf{r}}^j\right)$ is the accumulated transmittance. Similarly, depth $\hat{D}(\mathbf{r})$, normal $\hat{N}(\mathbf{r})$ of the surface can be accumulated as:
\begin{equation}\label{eq:depth}
    \hat{D}(\mathbf{r})=\sum_{i=1}^M T_{\mathbf{r}}^i \alpha_{\mathbf{r}}^i t_{\mathbf{r}}^i, \quad \hat{N}(\mathbf{r})=\sum_{i=1}^M T_{\mathbf{r}}^i \alpha_{\mathbf{r}}^i \hat{\mathbf{n}}_{\mathbf{r}}^i,
\end{equation}
where normal values $\mathbf{n}(\mathbf{x})$ can be estimated by computing the gradient of SDF function at point $\mathbf{x}$.





%% file: sections/method.tex

\input{sections/recon}

\input{sections/intrinsics}

%% file: sections/recon.tex
\section{Intrinsics Decomposition}\label{sec:recon}

\subsection{Geometry Field}
\subsubsection{Bubbling for Small Objects}\label{sec:bubble}


Unlike single-object scenarios, indoor scenes contain objects of different scales and different visibility levels, some of which appear rarely in the input views or are restricted in view poses due to their location (e.g. in the corner). We observe that existing indoor reconstruction methods~\cite{wang2022neuris,Yu2022MonoSDF}
frequently fail in recognizing and reconstructing thin (e.g. chair legs) or suspended (e.g. chandeliers) objects in the room, \emph{even with dense geometry priors}. \cref{fig:teaser} shows a reconstruction error on the chandelier.

The reason for the failure of recovering small objects in the scenes can be attributed to the inherent nature of the neural network as elaborated in ~\cite{tancik2020fourfeat}: the low-frequency information in a neural network tends to converge faster than the higher-frequency information. In an indoor scene, the SDF for large objects like walls and tables (corresponding to low frequency information) converges in earlier iterations. In some tasks the high-frequency details can be recovered in later iterations of training, while methods based on SDF as geometry representations can hardly recover the fine details due to \emph{vanishing gradients} for small objects.


\begin{figure}[h]
    \centering
    \hspace{-10pt}\includegraphics[width=230pt]{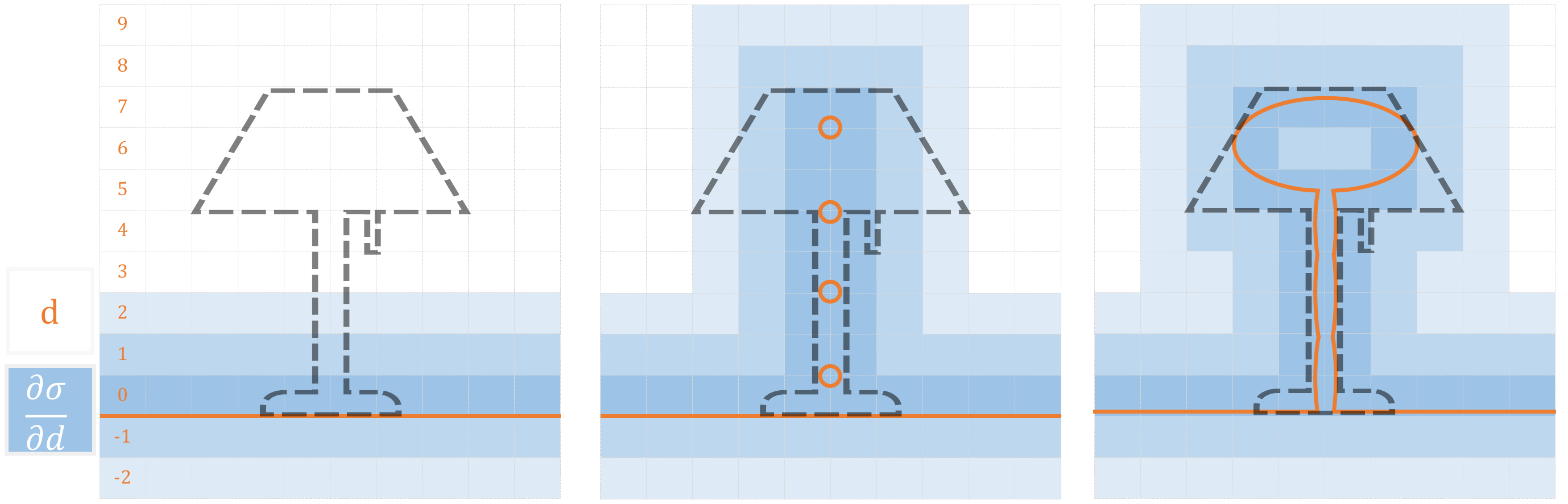}
    \caption{Concept image of bubble, where dotted lines represent thin geometry (lamp in this example) not yet learned by the SDF network. Left: $ \frac{\partial \sigma}{\partial d}$ rapidly vanishes as $d$ increases, and thus the SDF cannot learn the thin object. Middle: inserting bubbles (creating zero-value surfaces) recovers the gradient flow around the missing object. Right: the bubbles grow with the introduced gradients to recover the thin object. }
    \label{fig:vgrad}
\end{figure}

An example of the vanishing gradients is illustrated in ~\cref{fig:vgrad}. Suppose the loss for the radiance $\mathcal{L}$ is a function $c(\sigma(d(x;\theta)))$. The derivative of $\mathcal{L}$ w.r.t the network parameters $\theta$ for SDF can be written as $\frac{\partial \mathcal{L}}{\partial c} \frac{\partial c}{\partial \sigma} \frac{\partial \sigma}{\partial d}\frac{\partial d}{\partial \theta}$. For $d(x) \geq 0$, $ \frac{\partial \sigma}{\partial d} = \frac{1}{2 \beta^3} \exp \left(-\frac{d(\mathbf{x})}{\beta}\right)$ (from \cref{eq:density}). Usually the learned $\beta$ is large to make $\sigma$ fall off the target surface rapidly, indicating $ \frac{\partial \sigma}{\partial d}$ vanishes fast with increasing $d(x)$.
When the SDF for a surface with a thin object converges to the status as shown in the left of ~\cref{fig:vgrad}, the gradients  $ \frac{\partial \sigma}{\partial d}$  for points near the object off the surface are almost zero and therefore $\frac{\partial \mathcal{L}}{\partial \theta}$ remains near zero (see the gradient fields $ \frac{\partial \sigma}{\partial d}$ ~\cref{fig:vgrad}); in other words, no gradients can be acquired to recover the thin object. To address the problem, we propose to insert ``bubbles'' for the missing surface points to create gradients for SDF near small or thin objects. The middle figure of ~\cref{fig:vgrad} exemplifies inserted bubbles. The inserted bubble creates many local surface islands and gradient fields near them, which then allows the growing up of new objects that are previously ignored.  

Specifically, we obtain the bubbles from depth images. Given a depth image $D(u,v)$ with corresponding camera pose $[\mathrm{R}|\mathbf{t}]$ and intrinsics $\mathrm{K}$, the 3D point $\mathbf{x}(\mathbf{p})$ associated with a pixel $\mathbf{p}=(u,v)$ is
\begin{equation}\label{eq:unproj}
    \mathbf{x}(\mathbf{p},D) = \mathbf{t} + D(u,v) \left( \mathrm{R}\mathrm{K}^{-1} \left[ \begin{array}{c}
        u \\ v \\ 1
    \end{array} \right] \right).
\end{equation}


The 3D points from depth images, \ie the bubbles, can be regarded as an approximation of surface points during ray casting and its SDF value should be zero. In order to enforce precise surface reconstructions, we define bubble loss $\mathcal{L}_{\mathrm{bubble}}$ to minimize the absolute SDF value of these surface points:
\begin{equation}\label{eq:bubble}
    \mathcal{L}_{\mathrm{bubble}} = \sum_{\mathbf{p}\in\mathcal{P}}| d(\mathbf{x}(\mathbf{p},D)) |,
\end{equation}
where $d(\mathbf{x})$ is the predicted SDF value of a point $\mathbf{x}$ (\cref{eq:sdf}), $D$ is the depth image and $\mathcal{P}$ denotes the minibatch of sampled pixels in each iteration.

\subsubsection{Error-Guided Adaptive Sampling Strategy}\label{sec:adaptive}

Applying the bubble loss on the implicit SDF field can improve the 3D reconstruction quality. However, the geometry of large planar areas can already be reconstructed very well with the image-space depth loss. Applying $\mathcal{L}_{\mathrm{bubble}}$ on 3D points within these areas is a waste of computation. Furthermore, as analyzed in \cref{sec:intro}, small objects (which is our target) make up only a small percentage of all pixels, so it will be favorable if the bubble loss is applied more frequently in these areas.

Determining the areas of ``small objects'' is a perceptual task, which cannot be easily accomplished without semantic segmentation or manual marking. However, due to the feature of neural networks that low-frequency signals are always easier to fit than high-frequency signal~\cite{tancik2020fourfeat}, we can naturally filter out the uninterested large planar areas (low frequency) and preserve small-object areas (high frequency) according to the reconstruction error of the neural network. Specifically, given an error metric $E$ (depth loss is adopted in our case), we leverage \emph{importance sampling} algorithm according to a probability distribution determined by $E$. The PDF (Probability Density Function) of a point $\mathbf{x}(\mathbf{p};D)$ (\cref{eq:unproj}) is proportional to the reconstruction error $E(\mathbf{p})$ at the pixel $\mathbf{p}$. We also prune the pixels with errors lower than a threshold $P_\mathrm{min}$, indicating that those pixels are not of our interest. In this way, the network can pay more attention to the erroneous areas and converge faster. 
Please refer to our supplementary material for visualizations of PDF map.


The PDF values are updated dynamically: We maintain a PDF map for each training image. After each training iteration, the PDF values for pixels in the current batch are updated with their error metrics $E$.

\paragraph{Geometry loss.}
Together with the bubble loss, our geometry loss is as follows to approximate the geometry field:
\begin{align}\label{eq:traing}
    \mathcal{L}_\mathrm{geo} = &\lambda_1\mathcal{L}_\mathrm{eikonal} + \lambda_2\mathcal{L}_\mathrm{depth}  + \lambda_3\mathcal{L}_\mathrm{normal} \notag \\ 
    & + \lambda_4\mathcal{L}_\mathrm{smooth} + \lambda_5\mathcal{L}_{\mathrm{bubble}} .
\end{align}

As suggested by previous work~\cite{yariv2021volume}, we apply Eikonal term~\cite{gropp2020implicit} to regularize SDF values 
\begin{equation}
    \mathcal{L}_\mathrm{eikonal} = \sum_{\mathbf{x}\in\mathcal{X}}(\| \nabla d(\mathbf{x}) \|_2 - 1)^2,
\end{equation}
where $\mathcal{X}$ is the minibatch of 3D points uniformly sampled in 3D space and nearby surface.

We use depth and normal priors to supervise our network to handle shape-radiance ambiguity. We compute the surface depth and normal 
by \cref{eq:depth} and use $L_1$ loss for depth and angular $L_1$ loss for normal:
\begin{align}
    \mathcal{L}_\mathrm{depth} &= \sum_{\mathbf{r}\in\mathcal{R}}\| \hat{D}(\mathbf{r}) - D(\mathbf{r}) \|_1, \\
    \mathcal{L}_\mathrm{normal} &= \sum_{\mathbf{r}\in\mathcal{R}}\| 1 - \hat{N}(\mathbf{r})\cdot N(\mathbf{r}) \|_1.
\end{align}

We also use smoothness loss~\cite{Niemeyer2020CVPR} on the gradient of the SDF field to encourage smooth surface reconstruction: 
\begin{equation}\label{eq:smooth}
    \mathcal{L}_\mathrm{smooth} = \sum_{\mathbf{x}\in\mathcal{S}}\| \nabla d(\mathbf{x}) - \nabla d(\mathbf{x}+\epsilon) \|_2,
\end{equation}
where $\mathcal{S}$ is the minibatch of points sampled near the surface and $\epsilon$ is a small random uniform 3D perturbation.

\subsection{Emitter Semantic Field}\label{sec:emit}

Considering that our radiance field $F_c$ is trained from LDR images, the light intensity from emitters (e.g. lamps/windows) is usually under-estimated. This will cause an over-dark estimation in the re-render stage (\cref{sec:render}). We resolve this by introducing semantic labels of emitters to optimize the radiance value emitted from light sources. We add a neural emitter semantic field $F_e$ into our module, which determines whether the input 3D point $\mathbf{x}$ is on an emitter or not. The estimated emitter mask $\hat{M}_e(\mathbf{r})$ for a ray can also be evaluated by volume accumulation
\begin{align}
    \hat{m}(\mathbf{x}) &= F_e(\mathbf{z}(\mathbf{x})), \\
    \hat{M}_e(\mathbf{r}) &= \sum_{i=1}^M T_{\mathbf{r}}^i \alpha_{\mathbf{r}}^i \hat{m}_{\mathbf{r}}^i,
\end{align}
where $\mathbf{z}(\mathbf{x})$ is the latent code output by SDF field $F_d$ (\cref{eq:sdf}). We choose $\mathbf{z}(\mathbf{x})$ as the MLP input because it can provide latent information about the scene.

\paragraph{Emitter segmentation loss.} Given ground truth emitter masks $M_e$, we optimize $F_e$ by a binary cross-entropy loss:
\begin{equation}
    \mathcal{L}_\mathrm{emi} = \sum_{\mathbf{r}\in\mathcal{R}} M_e(\mathbf{r})\log{\hat{M}_e(\mathbf{r})} + (1 - M_e(\mathbf{r}))\log{(1 - \hat{M}_e(\mathbf{r}))}.
\end{equation}

After $F_e$ is trained, it can be used by our raytracing stage to indicate if a ray hits an emitter. We use K-Means algorithm~\cite{hartigan1979algorithm} to cluster emitter points as $K$ emitters. To model HDR emissions, we define an array $\mathbf{L}[\cdot]$ with size $K$ as a learnable parameter that corresponds to the emission values of each emitter. $\mathbf{L}[\cdot]$ will be queried in the raytracing and re-rendering stage (see \cref{sec:render}).

\subsection{Material Field}
Similar to the geometry and radiance field, we parameterize the spatially-varying material of the scene as a neural field. We use physically-based GGX microfacet BRDF model~\cite{walter2007microfacet} to present scene material and introduce two MLPs to model the albedo and roughness of the scene, respectively:
\begin{align}
    K_d(\mathbf{x}), K_s(\mathbf{x}) &= F_a(\mathbf{z}(\mathbf{x})), \\
    \rho(\mathbf{x}) &= F_\rho(\mathbf{z}(\mathbf{x})),
\end{align}
where $K_d(\mathbf{x})$ and $K_s(\mathbf{x})$ are the diffuse and specular albedo at 3D point $\mathbf{x}$, $\rho(\mathbf{x})$ is the surface roughness at $\mathbf{x}$. Note that the SDF field $F_d$ and radiance field $F_c$ have been pretrained and fixed. The estimated material parameter associated with a ray $\mathbf{r}$ can be calculated by volumetric accumulation similar to \cref{eq:rgb} and \cref{eq:depth}.

\paragraph{Material regularizations.} To enforce physical correctness for predicted material parameters, we define regularizations as
\begin{align}\label{eq:material}
    \mathcal{L}_\mathrm{mreg} = & \sum_{\mathbf{x}\in\mathcal{S}} {\| \hat{M}(\mathbf{x}) - \hat{M}(\mathbf{x} + \epsilon) \|_2}\\
    + & \sum_{\mathbf{x}\in\mathcal{S}} \left(\hat{K}_d(\mathbf{x}) + \hat{K}_s(\mathbf{x}) - 1\right)_+,
\end{align}
where $\hat{M}\in\{ \hat{K}_d, \hat{K}_s, \hat{\rho} \}$ denotes 3 material parameters. Similar to \cref{eq:smooth}, we encourage smooth estimation of materials in the first loss term. According to energy conservation law, the sum of diffuse and specular albedo should not exceed 1 (the second loss term), where $(\cdot)_+$ is the ReLU function~\cite{nair2010rectified}.

%% file: sections/intrinsics.tex
{
\begin{table*}[h]
    \centering
    \caption{\textbf{Quantitative comparisons of novel view synthesis results.} Data in brackets denote metrics in training views. }
    \vspace{-1ex}
    \begin{tabular}{ccccccc}\hline 
        \multirow{2}{*}{Method} & \multicolumn{3}{c}{Synthetic Data} & \multicolumn{3}{c}{Real Data} \\
         & PSNR$\uparrow$ & SSIM$\uparrow$ & LPIPS$\downarrow$ & PSNR$\uparrow$ & SSIM$\uparrow$ & LPIPS$\downarrow$ \\\hline 
        NeuRIS \cite{wang2022neuris} & 25.02 (26.01) & 0.77 (0.77) & 0.46 (0.42) & 22.33 (24.45) & 0.68 (0.71) & 0.53 (0.50) \\
        MonoSDF  \cite{Yu2022MonoSDF}& 25.74 (26.73) & 0.79 (0.78) & 0.38 (0.43) & 22.38 (24.71) & 0.69 (0.73) & 0.49 (0.45) \\\hline 
        NeRF \cite{mildenhall2020nerf} & 26.81 (27.74) & 0.85 (0.85) & 0.18 (0.22) & 24.06 (25.58) & 0.77 (0.78) & 0.27 (0.27) \\ 
        Instant-NGP \cite{mueller2022instant} & 23.89 (\textbf{31.59}) & 0.78 (\textbf{0.90}) & 0.25 (\textbf{0.12}) & 22.95 (\textbf{28.51}) & 0.76 (\textbf{0.88}) & \textbf{0.20} (\textbf{0.09}) \\\hline
        Ours & \textbf{28.42} (29.70) & \textbf{0.87} (0.87) & \textbf{0.15} (0.19) & \textbf{24.66} (26.33) & \textbf{0.78} (0.79) & 0.25 (0.26) \\\hline
    \end{tabular}
    \label{tab:novel}
\end{table*}
}
{
\setlength\tabcolsep{1.5pt}
\begin{figure*}[h]
    \centering
    \begin{tabular}{m{2.5cm}<{\centering}m{2.5cm}<{\centering}m{2.5cm}<{\centering}m{2.5cm}<{\centering}m{2.5cm}<{\centering}m{2.5cm}<{\centering}}
         GT & Ours & NeRF & Instant-NGP & NeuRIS & MonoSDF \\
         \stackinset{l}{1pt}{b}{1pt}{\includegraphics[width=1.7cm]{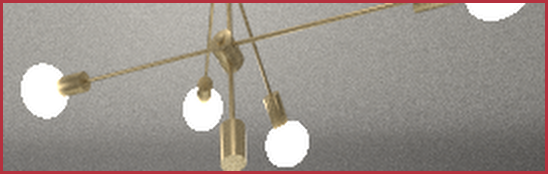}}{\includegraphics[width=2.5cm]{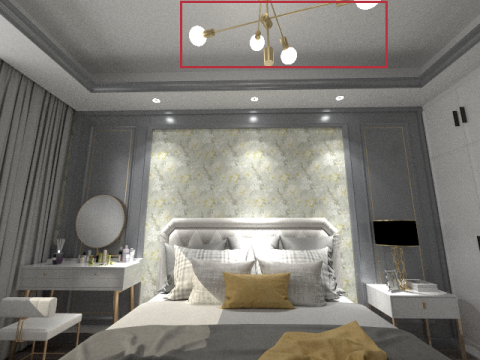}} & \stackinset{l}{1pt}{b}{1pt}{\includegraphics[width=1.7cm]{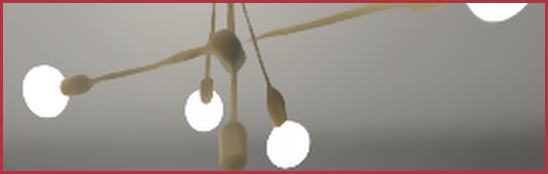}}{\includegraphics[width=2.5cm]{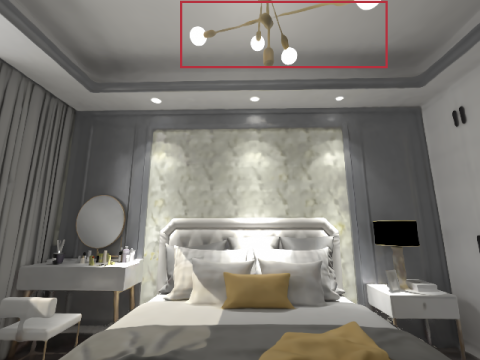}} & 
         \includegraphics[width=2.5cm]{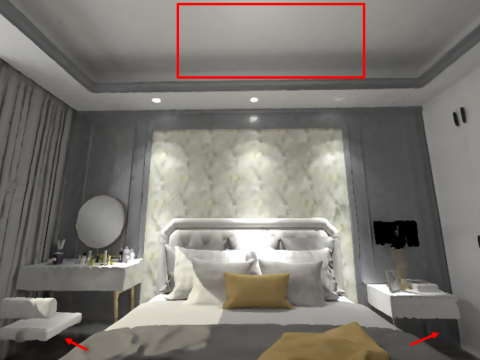} & \stackinset{l}{1pt}{b}{1pt}{\includegraphics[width=1.7cm]{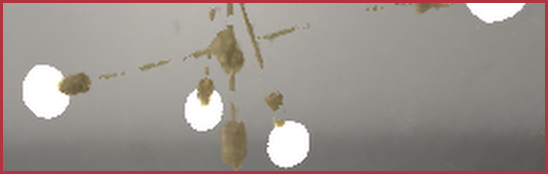}}{\includegraphics[width=2.5cm]{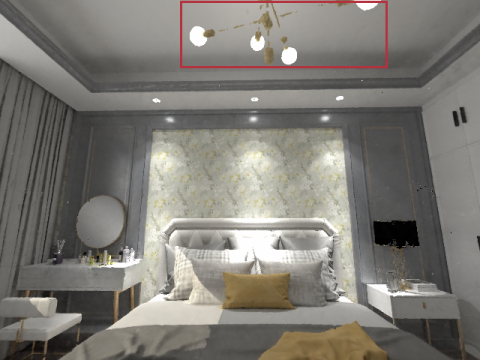}} 
         & \includegraphics[width=2.5cm]{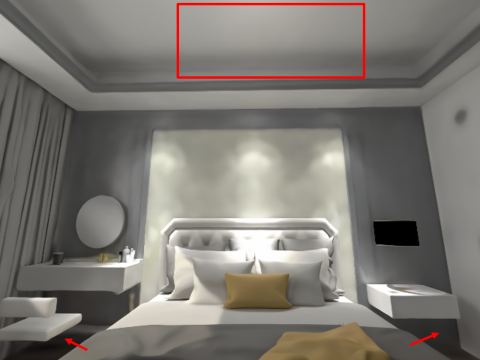} & \includegraphics[width=2.5cm]{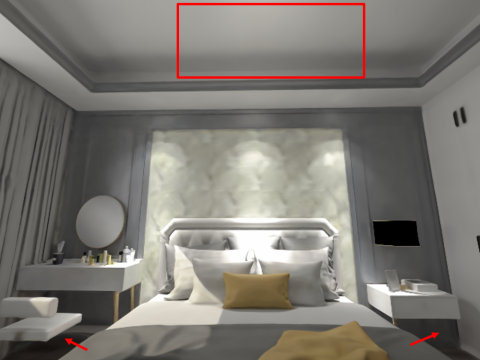} \\
         \includegraphics[width=2.5cm]{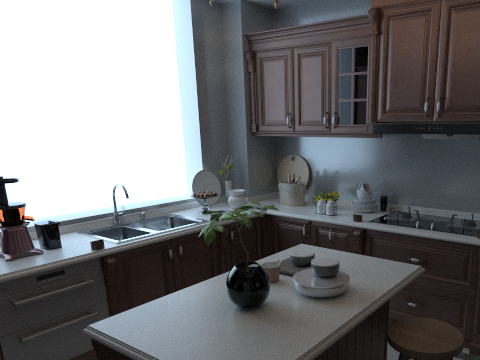} & \includegraphics[width=2.5cm]{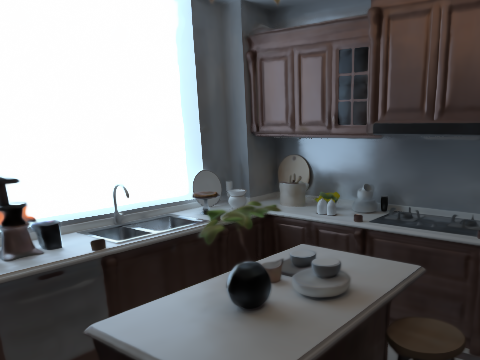} & \includegraphics[width=2.5cm]{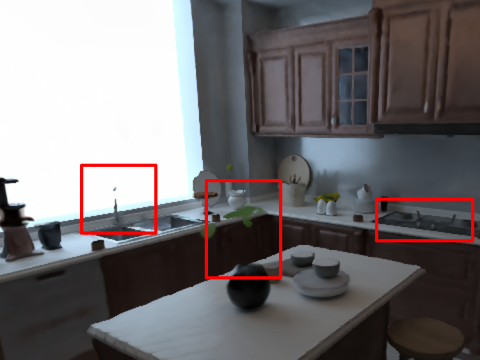} & \includegraphics[width=2.5cm]{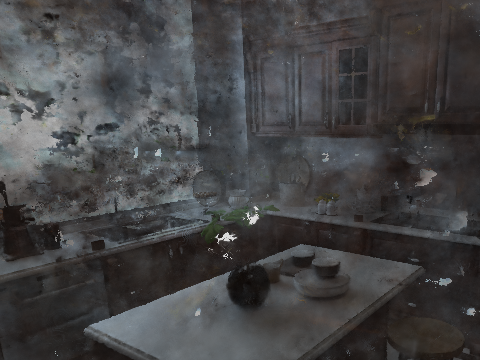} & \includegraphics[width=2.5cm]{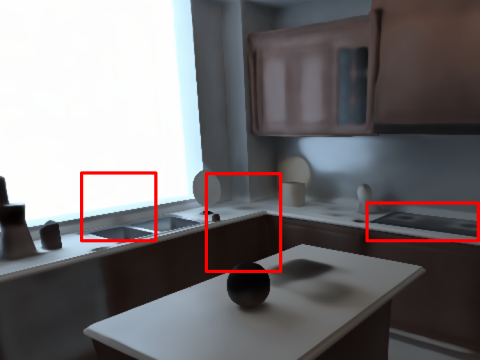} & \includegraphics[width=2.5cm]{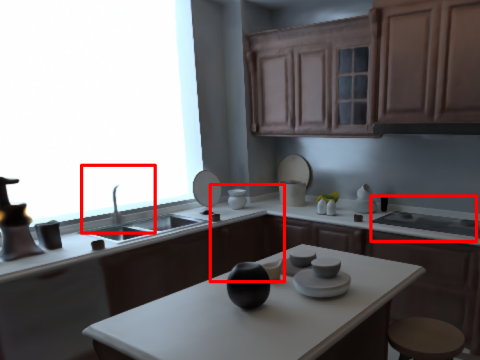} \\
    \end{tabular}
    \vspace{-1ex}
    \caption{\textbf{Qualitative comparisons of novel view synthesis.}  More results are presented in supplementary material.}
    \label{fig:novel}
    \vspace{-1ex}
\end{figure*}

}

\section{Differentiable Monte Carlo Raytracing}\label{sec:render}

Given material, geometry, and lighting components, the scene appearance can be re-rendered using surface rendering algorithms.  Inspired by~\cite{zhu2022learning}, we use differentiable Monte Carlo rendering algorithm with raytracing to recover scene appearance from shape, material and illumination. The difference between our work and~\cite{zhu2022learning} is that their raytracing is performed on screen space while ours is in a 3D volumetric space. For a ray $\mathbf{r}: \mathbf{x}=\mathbf{o}+t\mathbf{v}_\mathbf{s}$, we can trace and intersect it with the neural SDF field. The intersection point $\mathbf{s}$ can be estimated by
\begin{equation}\label{eq:ray}
    \mathbf{s} = \mathrm{trace}(\mathbf{r}) = \mathbf{o} + \left(\sum_{i=1}^M T_{\mathbf{r}}^i \alpha_{\mathbf{r}}^i t_{\mathbf{r}}^i\right)\mathbf{v}_\mathbf{s}.
\end{equation}

We leverage Monte Carlo rendering technique to perform the scene re-rendering. We first cast the rays from camera view to obtain the surface points associated with each pixel by \cref{eq:ray}, as well as their corresponding surface normal by \cref{eq:depth}. Then, given a sample rate $N$, we use GGX importance sampling to generate $N$ outgoing rays $\{\mathbf{r}_\mathbf{s}^k: \mathbf{x} = \mathbf{s} + t\mathbf{d}_\mathbf{s}^k\}_{k=1}^N$ starting from a surface point $\mathbf{s}$ according to the surface normal and material parameters $\hat{N}(\mathbf{s}), \hat{K}_d(\mathbf{s}),\hat{K}_s(\mathbf{s}),\hat{\rho}(\mathbf{s})$. The surface color can be rendered by Monte Carlo integration:
\begin{equation}
    \hat{\mathbf{R}}(\mathbf{s}) = \frac{1}{N}\sum_{k=1}^N{\frac{f_r(\mathbf{v}_\mathbf{s},\mathbf{d}_\mathbf{s}^k; \hat{N}, \hat{K}_d,\hat{K}_s,\hat{\rho})L_\mathbf{s}^k\cos{\theta_k}}{p(\mathbf{v}_\mathbf{s},\mathbf{d}_\mathbf{s}^k)}},
    \label{eq:render}
\end{equation}
where $f_r$ and $p$ is the evaluation and PDF value of GGX microfacet BRDF model determined by the material parameters. $L_\mathbf{s}^k$ is the predicted radiance of ray $\mathbf{r}_\mathbf{s}^k$:
\begin{equation}\label{eq:radiance}
        L_\mathbf{s}^k = \begin{cases}
        \hat{\mathbf{C}}(\mathbf{r}_\mathbf{s}^k) & \text{if not }\hat{M}_e(\mathbf{r}_\mathbf{s}^k), \\
        \mathbf{L}[\mathrm{index}(\mathrm{trace}(\mathbf{r}_\mathbf{s}^k))] & \text{if }\hat{M}_e(\mathbf{r}_\mathbf{s}^k),
    \end{cases}
\end{equation}
which can be divided into two cases: we use $F_e$ to determine if $\mathbf{r}_\mathbf{s}^k$ hits an emitter. If so, we obtain the emitter index by K-means and retrieve its emission from emission field $\mathbf{L}[\cdot]$ (defined in \cref{sec:emit}). Otherwise, we use $F_c$ and volume rendering (\cref{eq:rgb}) to predict the radiance of the ray.

\section{Training}\label{sec:train}
The training of intrinsic decomposition and the reconstruction of the indoor scenes are conducted in two stages: 1) the training of geometry and radiance fields, 2) the training of the material and emission fields.


\paragraph{Training of geometry and radiance fields.}\label{sec:train1}

The training scheme of the SDF network $F_d$ and radiance network $F_c$ is end-to-end. We firstly use geometric initialization~\cite{Atzmon_2020_CVPR} to initialize $F_d$, and then optimize the networks with following loss:
\begin{align}\label{eq:train1}
    \mathcal{L}_\mathrm{1} = \mathcal{L}_\mathrm{rgb} + \lambda_\mathrm{geo}\mathcal{L}_\mathrm{geo}  + \lambda_\mathrm{emi}\mathcal{L}_{\mathrm{emi}},
\end{align}
where $ \mathcal{L}_\mathrm{rgb} = \sum_{\mathbf{r}\in\mathcal{R}}\| \hat{C}(\mathbf{r}) - C(\mathbf{r}) \|_1$.  \cref{eq:rgb} provides volume rendering results from 3D neural representation to 2D images. $\mathcal{R}$ denotes the set of pixels/rays sampled in the minibatch and $C(\mathbf{r})$ is the ground truth pixel color.


The weight hyperparameters of some losses vary during our 3-step training:
\begin{enumerate}
    \item (Warm-up step) In the early stage of training, bubble loss is not applied (\ie $\lambda_5=0$) until the network can reconstruct a coarse scene geometry. At the end of this step, the error (PDF) maps are initialized per image.
    \item (Bubble step) The adaptive sampling and bubble loss are enabled in this step to reconstruct missing small structures. Note that the bubble loss breaks the stable status of the converged SDF field so far, which makes the Eikonal and smoothness regularization increase. Therefore, we disable them (\ie $\lambda_1,\lambda_4=0$) to prevent potential contradictions.
    \item (Smooth step) Since the bubble loss will affect the smoothness and stability of the SDF field, which will have negative effects such as imprecision in normal estimation. Therefore, the bubble loss is disabled (\ie $\lambda_5=0$) in this step. To restore the SDF field smoothness, the training continues with the Eikonal and smoothness loss enabled again.    
\end{enumerate}

\paragraph{Training of material and emission fields.}
We jointly train the material network $F_a$, $F_\rho$ and the emission array $\mathbf{L}[\cdot]$ after the geometry and radiance networks ($F_d,F_c,F_e$) have been pretrained. During the optimization of intrinsic decomposition, the parameters of the geometry and radiance networks are fixed. Since the ground truths of material are impossible to capture from images, we weakly supervise the network by re-render results instead of direct supervision from strong material priors. 
The overall loss for this stage is 
\begin{equation}
    \mathcal{L}_\mathrm{2} = \mathcal{L}_\mathrm{render} + \lambda_\mathrm{mat} \mathcal{L}_\mathrm{mat},
\end{equation}
where $\mathcal{L}_{\mathrm{render}} = \sum_{\mathbf{r}\in\mathcal{R}} \| \hat{\mathbf{R}}(\mathbf{s}(\mathbf{r})) - \mathbf{C}(\mathbf{r}) \|_1$,  minimizing the $L_1$ error between the re-rendered result (\cref{eq:render}) and input image.

%% file: sections/experiment.tex
\section{Experiments}

We first analyze and compare our method with the state-of-the-art methods in terms of geometry reconstruction and novel view synthesis. We then demonstrate qualitative scene editing and relighting results based on our intrinsics decomposition method. Finally, we also perform ablation studies to prove the effectiveness of our design.

\paragraph{Datasets.} We propose a new synthetic multi-view indoor scene dataset, 
which includes well-designed scenes by artists and provides high quality rendered images with ground truth camera poses and geometry annotations (depth and normal maps). Existing datasets (such as ScanNet~\cite{dai2017scannet}) suffers from inaccurate camera calibration, erroneous depth capture and low image quality (such as motion blur), which will crucially affect the reconstruction quality. Our dataset provides well-designed indoor scenes with ground truth camera poses, normal and depth maps, with superior image quality to existing datasets. We also test our method on real-world scenes, specifically, a living room scene from \cite{PMGD21} and 3 scenes from Scalable-NISR~\cite{wu2022snisr}. All the real data contains calibrated camera poses and depth maps. To train the emitter field, we annotate emitter masks for the real data.

\paragraph{Metrics.} For the 3D geometry reconstruction, following previous works~\cite{Yu2022MonoSDF}, we compare on mesh-based metrics including accuracy, precision, recall and F-score. We also present image-space geometry errors including depth error and normal angular error in the supplementary. For the novel view synthesis, we use widely-used image metrics including PSNR, SSIM~\cite{wang2004image} and LPIPS~\cite{zhang2018perceptual}.

\paragraph{Baselines.} We compare against state-of-the-art neural reconstruction, multi-view stereo and novel view synthesis methods. In particular, for the geometry reconstruction, VolSDF~\cite{yariv2021volume}, NeuRIS~\cite{wang2022neuris} and MonoSDF~\cite{Yu2022MonoSDF} are compared. Since the original VolSDF method suffers from shape-radiance ambiguity and usually fails in reconstructing plausible scene structure, we add an additional depth loss when optimizing it. We mark it as \emph{``VolSDF-D''} in the following figures and tables. For novel view synthesis, we also compare with NeRF~\cite{mildenhall2020nerf} and Instant-NGP~\cite{mueller2022instant}.



\subsection{Comparisons with state-of-the-art methods}

\paragraph{Novel view synthesis.} We evaluate the quality of novel view synthesis on both synthetic and real data. \cref{tab:novel} and \cref{fig:novel} give quantitative results and qualitative results. NeRF and SDF-baselines struggle in recognizing small objects, leading to poor results. Instant-NGP performs best in training views. However, this is achieved by color over-fitting instead of accurate geometry understanding, leading to poor quality in test views. Benefiting from our high-quality geometry, our method provides good results in novel views. We also provide view interpolation results, which are displayed in the supplementary video. 

{
\setlength\tabcolsep{5pt}
\begin{table}[h!]
    \centering
    \caption{\textbf{Quantitative comparisons of geometric reconstruction results} on synthetic data.}
    \vspace{-0.5ex}
    \begin{tabular}{ccccc}\hline
        Method & Ours & VolSDF-D & NeuRIS & MonoSDF \\\hline
        Acc.$\downarrow$ & \textbf{0.035} & 0.041 & 0.036 & 0.052 \\
        Prec.$\uparrow$ & \textbf{0.87} & 0.76 & 0.74 & 0.82 \\
        Recall$\uparrow$ & \textbf{0.79} & 0.64 & 0.65 & 0.72 \\
        F-Score$\uparrow$ & \textbf{0.83} & 0.68 & 0.66 & 0.77 \\\hline
    \end{tabular}
    \label{tab:mesh}
    \vspace{-2ex}
\end{table}
}

\paragraph{Geometry reconstruction.} We evaluate 3D geometry metrics on our synthetic dataset with ground truth meshes. \cref{tab:mesh} shows quantitative results on mesh evaluation. Our method outperforms all baselines due to the precise reconstructions on small objects. \cref{fig:geo} shows qualitative results on the reconstructed depth and normal maps. Our method can faithfully reconstruct thin structures where baseline fails.
Please refer to our supplementary material for 
more qualitative and quantitative results.

{
\setlength\tabcolsep{2pt}
\begin{figure}[h!]
    \centering
    \begin{tabular}{m{2.7cm}<{\centering}m{2.7cm}<{\centering}m{2.7cm}<{\centering}}
        {Image} & {Ours} & {MonoSDF~\cite{Yu2022MonoSDF}} \\
        \includegraphics[width=2.7cm]{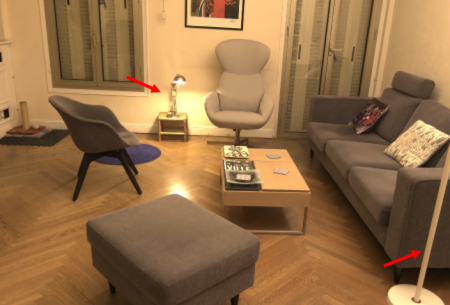} & \includegraphics[width=2.7cm]{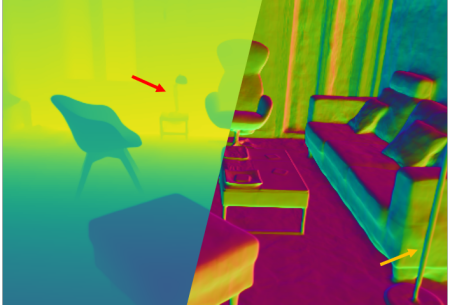} & \includegraphics[width=2.7cm]{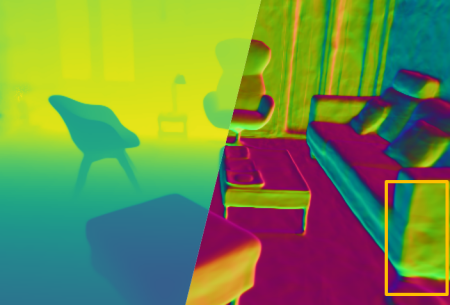} \\
    \end{tabular}
    \vspace{-1ex}
    \caption{\textbf{Qualitative comparisons of reconstructed depth map and normal map (real data).} Zoom in for details.}
    \label{fig:geo}
    \vspace{-1ex}
\end{figure}
}





\subsection{Scene Editing} 

With the decomposition results of shape, material and lighting, we can enable photo-realistic scene editing tasks such as material editing and relighting, as shown in \cref{fig:teaser,fig:edit}. In Fig.~\ref{fig:edit}, we change the hue of the light in the room  ($1^\text{{st}}$ column), increase the emission intensity of the lamp ($2^\text{{nd}}$ column), and change the material of the closet door into a mirror ($3^\text{{rd}}$ column), respectively. Note that the reflections in the specular mirror is consistent to the surroundings. With our physically-based rendering algorithm, our method can produce photo-realistic lighting effects such as specular reflections. More results and videos are presented in the supplementary.

{
\setlength\tabcolsep{1pt}
\begin{figure}[h]
    \centering
    \begin{tabular}{ccc}
         \includegraphics[height=1.8cm]{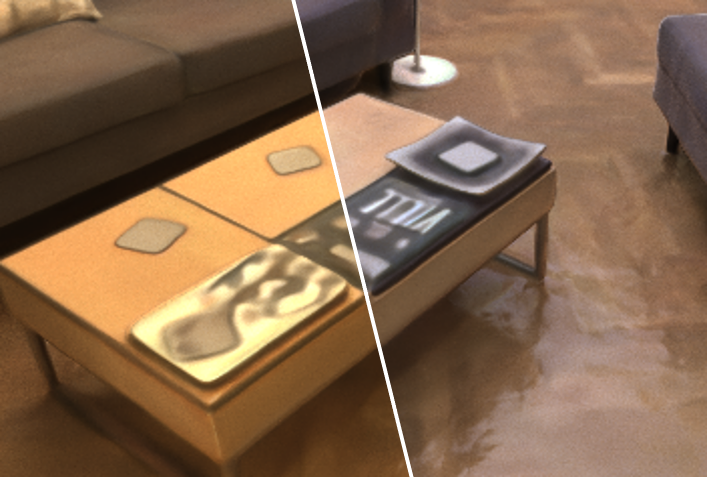} & \includegraphics[height=1.8cm]{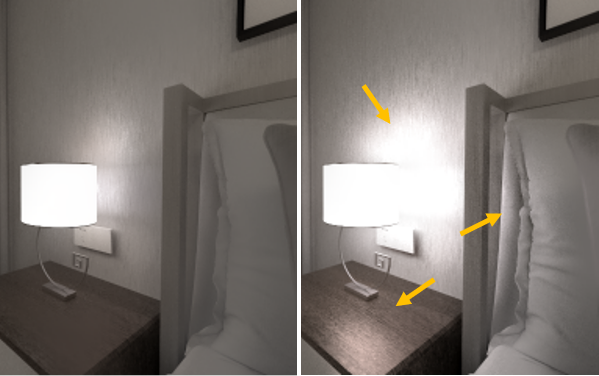} & \includegraphics[height=1.8cm]{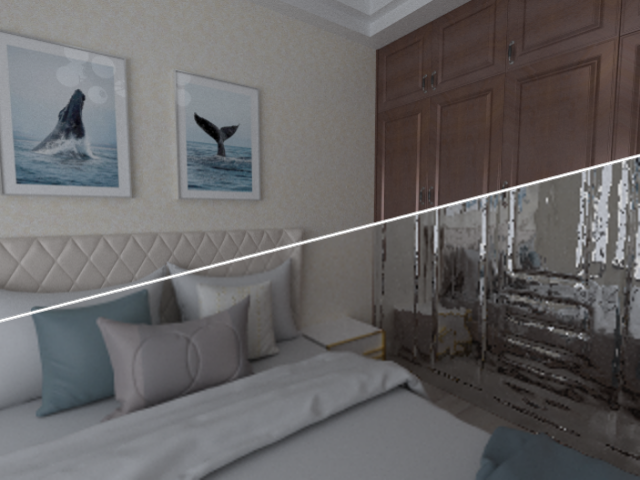}
    \end{tabular}
    \vspace{-0.5ex}
    \caption{\textbf{Qualitative results in material editing and relighting.}}
    \label{fig:edit}
\end{figure}
}



\subsection{Ablation Studies}



\begin{figure}[h!]
    \centering
    \vspace{-2ex}
    \begin{minipage}{0.45\linewidth}
        \begin{tabular}{cc}
         w/o noise & w/ noise \\
         \includegraphics[height=1.9cm]{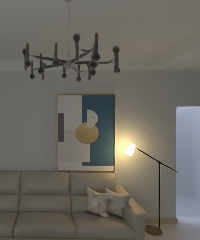} & \includegraphics[height=1.9cm]{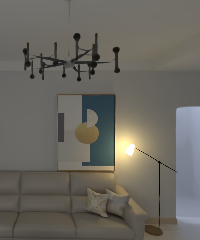}
        \end{tabular}
        \caption{Noisy depth.}
        \label{fig:ab_noise}
    \end{minipage}%
    \begin{minipage}{0.45\linewidth}
        \begin{tabular}{cc}
         Adaptive & Uniform \\
         \includegraphics[width=1.9cm]{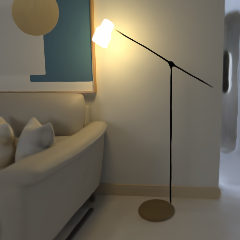} & \includegraphics[width=1.9cm]{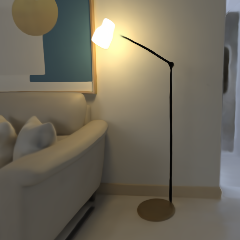}
        \end{tabular}
        \caption{Sampling strategy.}
        \label{fig:ab_sample}
    \end{minipage}
    \vspace{-1ex}
\end{figure}
\begin{table}[h!]
    \centering
    \caption{\textbf{Ablation studies on noisy depth and sampling strategy.} ``$\sigma$ noise'' and ``$3\sigma$ noise'' means using the standard noise model and 3 times noise model. ``$\sigma$ noise'' induces negligible negative effects, while ``$3\sigma$ noise'' still outperforms ``Uniform''.}
    \vspace{-2ex}
    \begin{tabular}{ccccc}\hline
        Method & Ours & $\sigma$ noise & $3\sigma$ noise & Uniform  \\\hline
        PSNR & 29.42 & 29.30 & 28.94 & 28.92 \\
        Depth-$L_1$ & {0.0326} & {0.0312} & 0.0328 & 0.0337 \\\hline
    \end{tabular}
    \label{tab:ablation}
    \vspace{-2ex}
\end{table}

\paragraph{Robustness on inaccurate depth information.} In real-world scenarios, the captured depth always contains noises and errors. To simulate this, we use noisy depths, which are added with noise scaling approximately quadratically with depth $z$, to supervise our bubble and depth loss. As suggested by~\cite{nguyen2012modeling,keselman2017intel}, the noise model is $\epsilon=N(\mu(z),\sigma(z))$, where $\mu(z)=0.0001125z^2+0.0048875,\sigma(z)=0.002925z^2+0.003325$. 
\cref{fig:ab_noise,tab:ablation} shows that noisy depths produce negligible impacts on the result, demonstrating the robustness of our method.




{
\paragraph{Effectiveness of adaptive sampling strategy.} We ablate between the sampling strategies on bubble points sampling. 
\setlength{\columnsep}{10pt}
We compare our error-guided adaptive sampling with uniform sampling. In \cref{fig:ab_sample}, uniform sampling is unable to reconstruct the complete lamp pole, due to insufficient bubble points sampled from the missing pole. In \cref{tab:ablation}, uniform sampling with ground truth depths is inferior to adaptive sampling with noisy depths.

}

%% file: sections/conclusion.tex
\section{Conclusion and Limitations}
This work proposes I$^2$-SDF that reconstructs an intrinsic neural scene from multi-view images, enabling physically-realistic novel view synthesis of editable indoor scenes. With the novel bubbling strategy, we are able to recover the small objects in large-scale scenes and obtain SOTA geometry and novel view synthesis results. This work's limitations are: Firstly, the MLP-based network backbone is not powerful enough to capture high-frequency textures. Secondly, the time-consuming MC raytracing increases the total reconstruction time. We consider them as problems to be solved in the future.

\paragraph{Acknowledgement.}
This work was supported in part by Key R\&D Program of Zhejiang Province (No. 2023C01039),  and NSFC (No. 61872319, No. 62233013), PI funding of Zhejiang Lab (121005-PI2101), Key Research Project of Zhejiang Lab (No. 2022PG1BB01).

%% file: sections/supp.tex
\section{Network Details}\label{sec:detail}

\subsection{Architecture}

All neural fields in our network are implemented by multi-layer perceptrons (MLPs). For neural SDF field $F_d$ and radiance field $F_c$, we follow VolSDF~\cite{yariv2021volume}'s default setting, where $F_d$ is a 8-layer MLP with hidden dimension 256 and $F_c$ is a 4-layer MLP with hidden dimension 256. The dimension of the latent code output by $F_d$ (\ie $\mathbf{z}(\mathbf{x})$) is 256, and a skip connection is used in the 4$^\text{{th}}$ layer in $F_d$. The input position $\mathbf{x}$ and view direction $\mathbf{v}$ are encoded by sinusoidal positional encoding~\cite{mildenhall2020nerf}, with a maximum frequency band of 6 for $\mathbf{x}$ and 4 for $\mathbf{v}$, the same as in VolSDF~\cite{yariv2021volume}. The emitter semantic field $F_e$ is a 2-layer MLP with hidden dimension 128, while material fields $F_\rho,F_a$ are 3-layer MLPs with hidden dimension 64.

\subsection{Implementation Details}

The network model, as well as the training and evaluation scripts, are implemented with Pytorch~\cite{paszke2019pytorch}. The network is trained per-scene on a single NVIDIA Tesla V100 GPU. We adopt two stage training scheme, the training details of the 2 stages are as follows:

\paragraph{Training of geometry and radiance fields.} We jointly optimize SDF network $F_d$, radiance network $F_c$ and emitter semantic network $F_e$ in this stage. We optimize our model for 200k iterations in this stage, which takes about 15 hours for a scene. The training loss
\begin{align}
    \mathcal{L}_\mathrm{geo} = &\lambda_1\mathcal{L}_\mathrm{eikonal} + \lambda_2\mathcal{L}_\mathrm{depth}  + \lambda_3\mathcal{L}_\mathrm{normal} \notag \\ 
    & + \lambda_4\mathcal{L}_\mathrm{smooth} + \lambda_5\mathcal{L}_{\mathrm{bubble}} \\
    \mathcal{L}_\mathrm{1} =& \mathcal{L}_\mathrm{rgb} + \lambda_\mathrm{geo}\mathcal{L}_\mathrm{geo}  + \lambda_\mathrm{emi}\mathcal{L}_{\mathrm{emi}}
\end{align}
where the weight hyperparameters are $\lambda_\mathrm{geo}=1$, $\lambda_\mathrm{emi}=0.5$, and $\lambda_2=0.1,\lambda_3=0.05$, respectively. For $\lambda_1,\lambda_4,\lambda_5$, since the training process is further divided into 3 steps (\ie warm-up, bubble and smooth), their values are adjusted during the training accordingly:
\begin{enumerate}
    \item In warm-up step, $\lambda_5=0$, $\lambda_1=0.1$, $\lambda_4=0$.
    \item In bubble step, $\lambda_1=\lambda_4=0$, $\lambda_5=0.5$.
    \item In smooth step, $\lambda_5=0$, $\lambda_1=0.1$, $\lambda_4=0.01$.
\end{enumerate}
The number of iterations assigned to the three steps are 50k, 100k and 50k in sequence.

We use error-guided adaptive sampling in bubble step, where the pruning threshold $P_\mathrm{min}=0.05$.

\paragraph{Training of material and emission fields.} In this stage, we use importance sampling and Monte Carlo estimation to compute the rendering result.
We generate $N$ outgoing rays to perform Monte Carlo integration. In practice, the sample rate $N$ is set to 16, which is a trade-off between quality and performance.

We jointly optimize $F_a,F_\rho,\mathbf{L}[\cdot]$ for 100k iterations. The bottleneck of computational cost lies in the prediction of incident radiance $L_\mathbf{s}^k$, which grows proportional to the sample rate $N$. With $N=16$, the training lasts for about 2-3 days.

{
\setlength{\tabcolsep}{2pt}
\begin{figure}[ht!]
    \centering
    \begin{tabular}{ccc}
        \includegraphics[width=2.6cm]{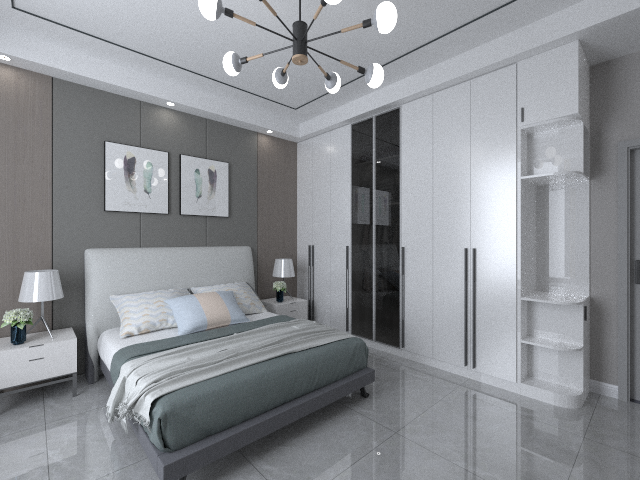} & \includegraphics[width=2.6cm]{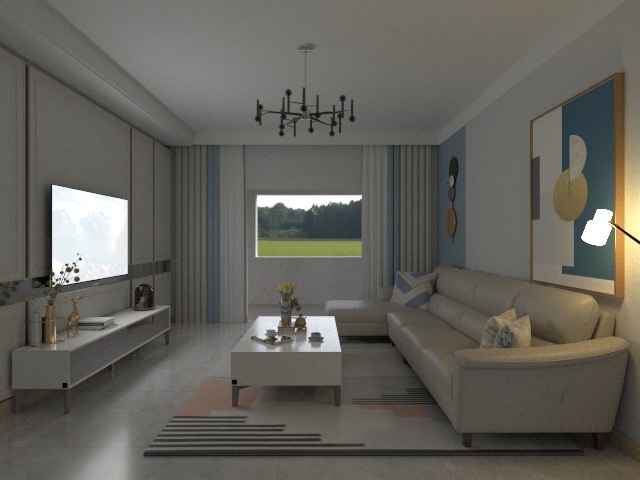} & \includegraphics[width=2.6cm]{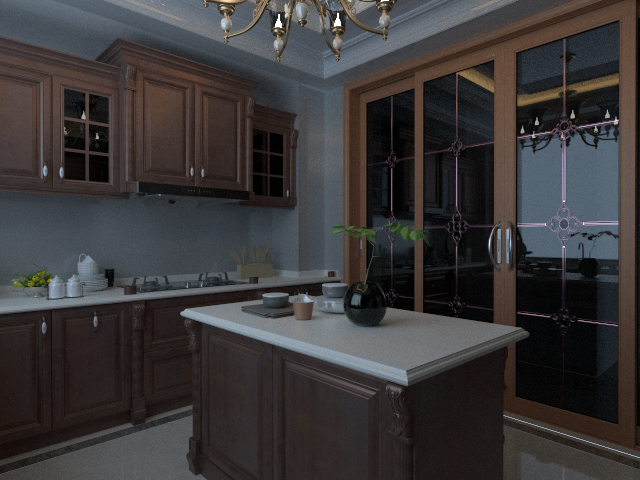} \\
        \includegraphics[width=2.6cm]{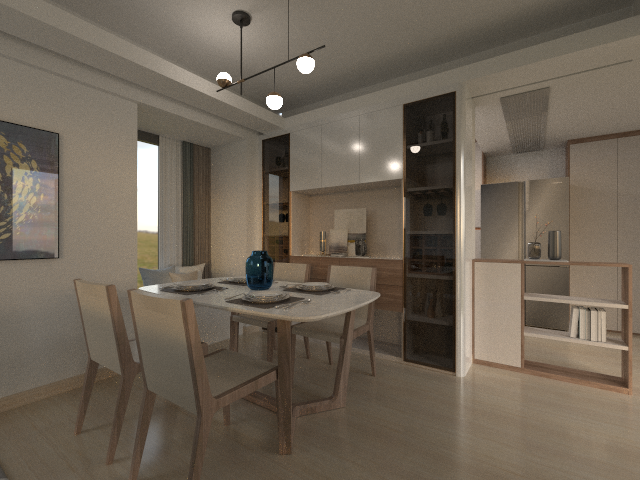} & \includegraphics[width=2.6cm]{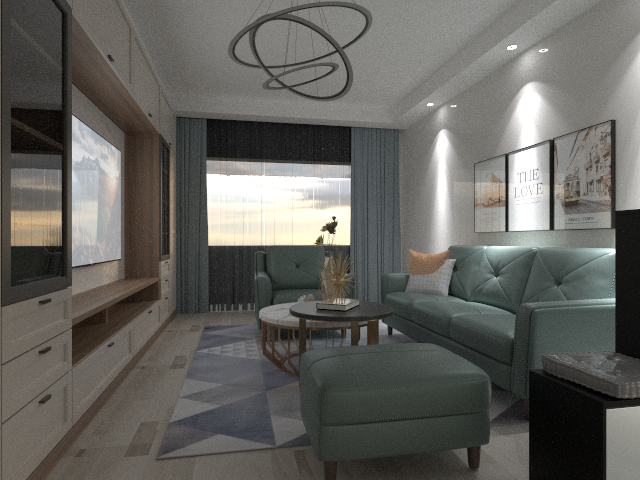} & \includegraphics[width=2.6cm]{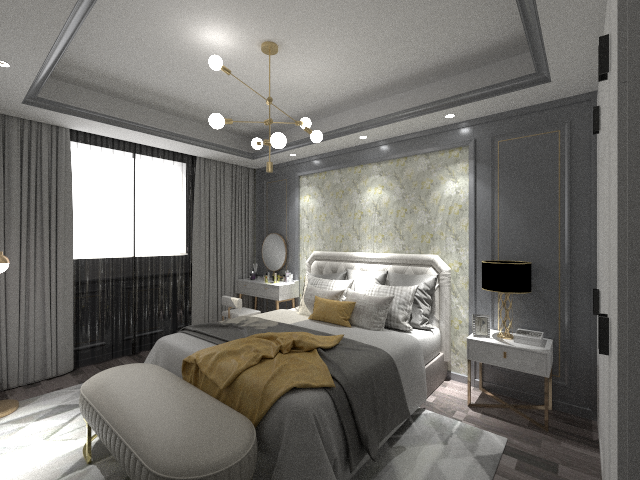}
    \end{tabular}
    \caption{Display of indoor scenes in our dataset.}
    \label{fig:dataset}
\end{figure}
}

\section{Dataset Details}\label{sec:dataset}

Our synthetic dataset contains 12 scenes in total: 6 bedrooms, 2 living rooms, 2 dining rooms and 2 kitchens. All the scenes are well-designed by artists with detailed geometry and fine textures. We use GPU-accelerated path tracing algorithm~\cite{kajiya1986rendering} to render the images, which can create photo-realistic rendering results with global illumination. All data are rendered on a NVIDIA RTX 3090 GPU, with 4096 samples per pixel (spp). The rendering time is roughly 15 seconds per image. \cref{fig:dataset} displays some of the indoor scenes in our dataset.

All images in our dataset are annotated by ground truth camera intrinsics and poses, normal maps, depth maps and emitter semantic masks.

{
\setlength\tabcolsep{1.5pt}
\begin{figure*}[ht!]
    \centering
    \begin{tabular}{ccccc}
        GT & 50k & 55k & 70k & 150k \\
        \includegraphics[width=2.7cm]{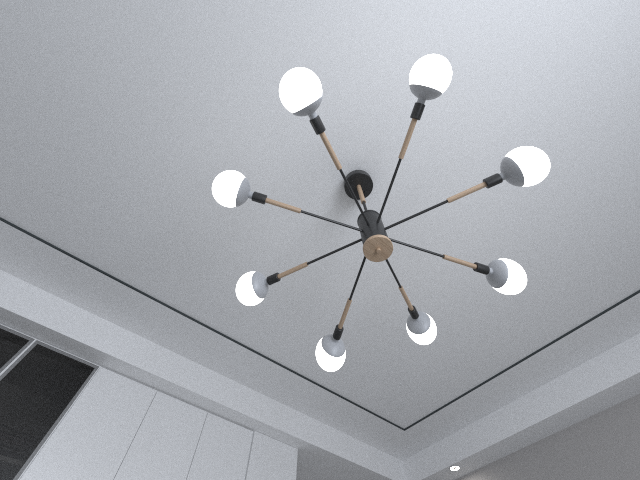} & \includegraphics[width=2.7cm]{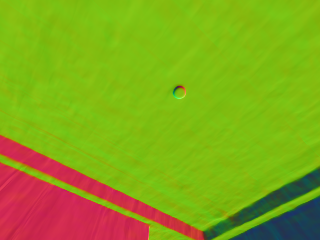} & \includegraphics[width=2.7cm]{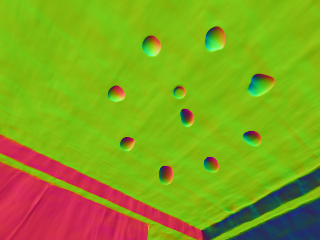} & \includegraphics[width=2.7cm]{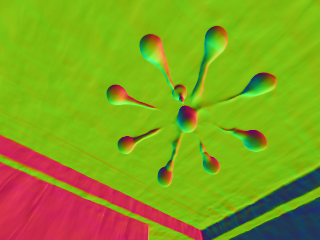} & \includegraphics[width=2.7cm]{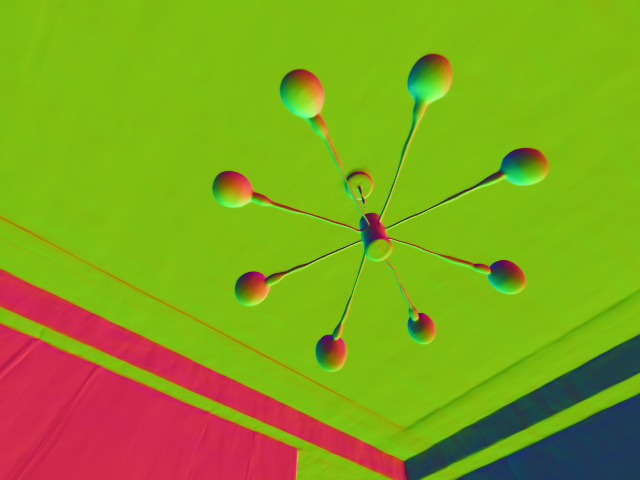}
    \end{tabular}
    \caption{\textbf{Intermediate results in bubble step.}}
    \label{fig:bubble}
\end{figure*}
}

\section{BRDF Model}

We use GGX microfacet BRDF model~\cite{walter2007microfacet} to approximate the surface reflection properties by a set of material parameters, including diffuse albedo $K_d$, specular albedo $K_s$ and roughness $\rho$. In our implementation, we refer to Unreal Engine~\cite{karis2013real}'s implementation of microfacet BRDF model. The BRDF (bidirectional reflectance distribution function) $f_r(\mathbf{v},\mathbf{d};N,K_d,K_s,\rho)$ (where $\mathbf{v}$ and $\mathbf{d}$ are view and lighting directions, and $N$ is the surface normal) can be decomposed into diffuse and specular components and computed by
\begin{align}
    f_r(\mathbf{v},\mathbf{d}) &= f_d(K_d)+f_s(\mathbf{v},\mathbf{d};N,K_s,\rho) \\
    f_d(K_d) &= \frac{K_d}{\pi}\quad\alpha = \rho^2 \\
    f_s(\mathbf{v},\mathbf{d};N,K_s,\rho) &= D(\alpha,N,h)G_2(\alpha,N,\mathbf{v},\mathbf{d})F(K_s,\mathbf{d},h)
\end{align}
where $f_d$ and $f_s$ are the diffuse and specular components respectively, while $D,G_2,F$ are the distribution, Fresnel and geometric terms, defined as
\newcommand{\nov}{(\mathbf{N}\cdot\mathbf{v})}
\newcommand{\nod}{(\mathbf{N}\cdot\mathbf{d})}
\begin{align}
    D(\alpha,\mathbf{N},\mathbf{h}) &= \frac{\alpha^2}{\pi((\alpha^2-1)(\mathbf{N}\cdot\mathbf{h})^2+1)^2} \\
    S(\alpha,\mathbf{N},\mathbf{v},\mathbf{d}) &= \nod\sqrt{\alpha^2+\nov^2(1-\alpha^2)} \\
    G_2(\alpha,\mathbf{N},\mathbf{v},\mathbf{d}) &= \frac{1}{2(S(\alpha,\mathbf{N},\mathbf{v},\mathbf{d}) + S(\alpha,\mathbf{N},\mathbf{d},\mathbf{v}))} \\
    \mathrm{lum}(C) &= 0.213C.\mathrm{r} + 0.715C.\mathrm{g} + 0.072C.\mathrm{b} \\
    F_{90}(K_s) &= \min(\frac{\mathrm{lum}(K_s)}{0.04},1) \\
    F(K_s,\mathbf{N},\mathbf{d}) &= K_s + (F_{90}(K_s)-K_s)(1 - \nod)^5
\end{align}

In importance sampling and Monte Carlo integration, we also need to calculate the PDF value $p(\mathbf{v},\mathbf{d})$ corresponding to the view and lighting direction:
\begin{align}
    w_d &= \frac{\mathrm{lum}(K_d)}{\mathrm{lum}(K_d)+\mathrm{lum}(K_s)} \\
    p(\mathbf{v},\mathbf{d}) &= w_d p_d(\mathbf{v},\mathbf{d}) + (1-w_d)p_s(\mathbf{v},\mathbf{d}) \\
    p_d(\mathbf{v},\mathbf{d}) &= \frac{N\cdot\mathbf{d}}{\pi} \\
    G_1(\alpha,N,\mathbf{v}) &= \frac{2}{\sqrt{1+\frac{\alpha^2(1-\nov^2)}{\nov^2}}+1} \\
    p_s(\mathbf{v},\mathbf{d}) &= \frac{D(\alpha,N,h)G_1(\alpha,N,\mathbf{v})}{4\nov}
\end{align}
where $p_d$ and $p_s$ are the diffuse and specular components, which are mixed according to the luminance of $K_d$ and $K_s$.


\section{Details of Bubbling and Adaptive Sampling}

\paragraph{Intermediate results in bubble step.} \cref{fig:bubble} presents the process of how the missing objects are reconstructed by our bubbling method. The chandelier is missing initially at 50k iterations. In the early stage of bubble step, the light balls are reconstructed rapidly (55k), since they are relatively large inside the chandelier. On the other hand, thin components (\eg poles) grow slowly (70k). Eventually (150k) the entire chandelier is successfully reconstructed.

{
\setlength\tabcolsep{1.5pt}
\begin{figure}[ht!]
    \centering
    \begin{tabular}{ccc}
        GT & 50k & 80k \\
        \includegraphics[width=2.7cm]{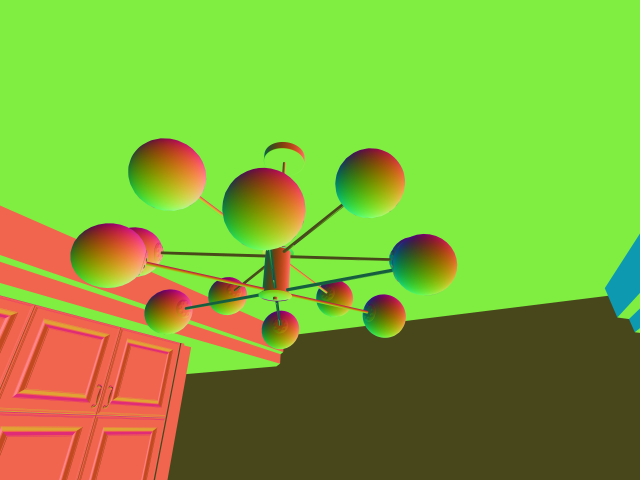} & \includegraphics[width=2.7cm]{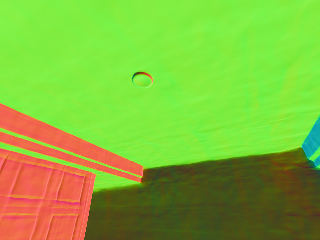} & \includegraphics[width=2.7cm]{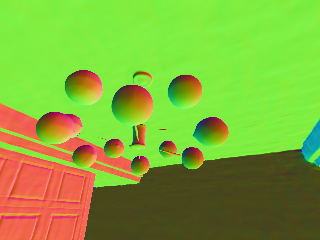} \\
        \includegraphics[width=2.7cm]{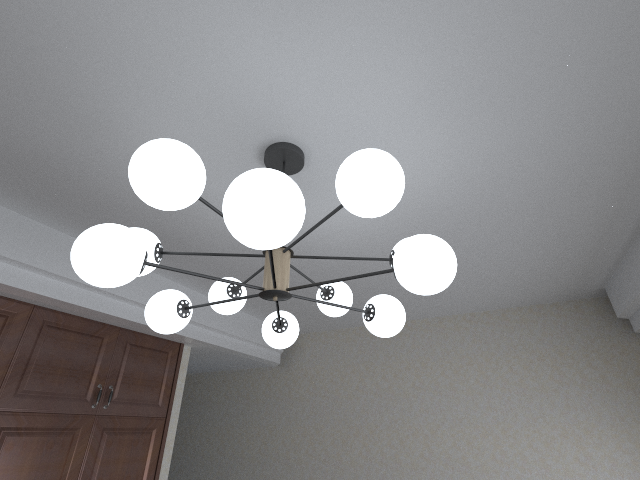} & \includegraphics[width=2.7cm]{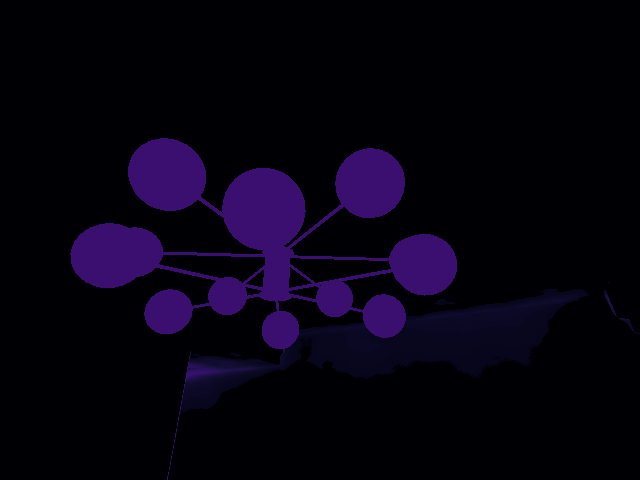} & \includegraphics[width=2.7cm]{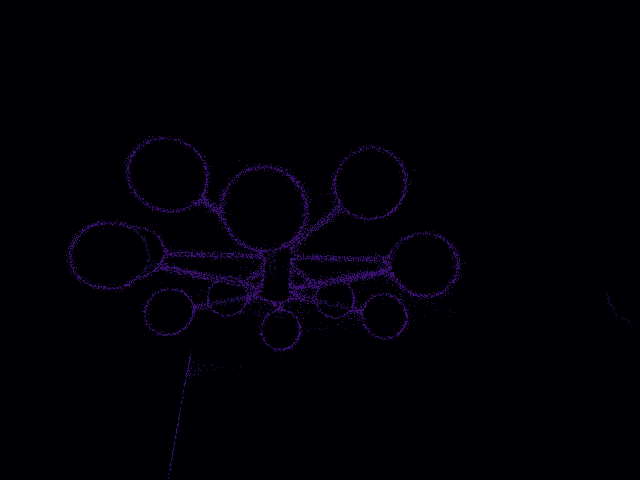}
    \end{tabular}
    \caption{\textbf{Visualization of error-guided sampling map during training.} The first row displays intermediate normal reconstruction results, while the second row displays the corresponding error PDF map.}
    \label{fig:adap}
\end{figure}
}

\paragraph{Visualization of error-guided sampling map during training.} As shown in \cref{fig:adap}, at the training iteration of 50k (\ie the start of bubble step), the chandelier is completely ignored and thus the corresponding pixels in the PDF map have high values. As the training proceeds to 80k iterations, the light balls have already been well-reconstructed, with chandelier poles still missing. Therefore, the value of pixels corresponding to light balls are reduced to 0, whereas chandelier pole pixels still need to be further sampled.

\begin{figure}[ht!]
    \setlength{\tabcolsep}{3pt}
    \centering
    \begin{tabular}{m{1em}<{\centering}m{2.4cm}<{\centering}m{2.4cm}<{\centering}m{2.4cm}<{\centering}}
        & $K_d$ & $K_s$ & $\rho$ \\
        \rotatebox{90}{GT} & \includegraphics[width=2.4cm]{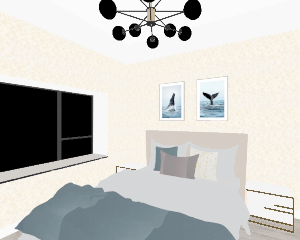} & \includegraphics[width=2.4cm]{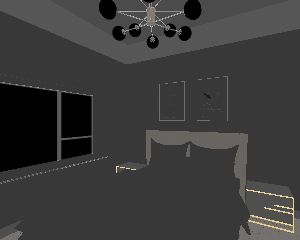} & \includegraphics[width=2.4cm]{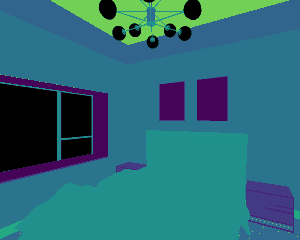} \\
        \rotatebox{90}{Pred} & \includegraphics[width=2.4cm]{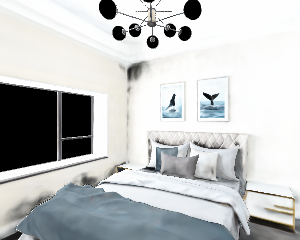} & \includegraphics[width=2.4cm]{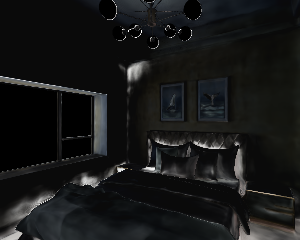} & \includegraphics[width=2.4cm]{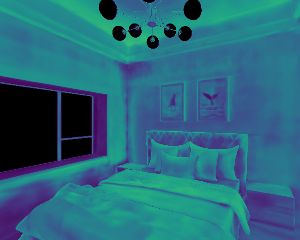}
    \end{tabular}
    \caption{\textbf{Comparisons between GT material and predictions.} Note that our optimization does not require supervision from GT material dataset 
    but we still produce plausible results.}
    \label{fig:mat}
\end{figure}

{
\setlength\tabcolsep{2pt}
\begin{figure*}[ht!]
    \vspace{-2ex}
    \centering
    \begin{tabular}{m{2.9cm}<{\centering}m{2.9cm}<{\centering}m{2.9cm}<{\centering}m{2.9cm}<{\centering}m{2.9cm}<{\centering}}
        {Image} & {Ours} & {NeuRIS~\cite{wang2022neuris}} & {MonoSDF~\cite{Yu2022MonoSDF}} & VolSDF-D \\
        \includegraphics[width=2.9cm]{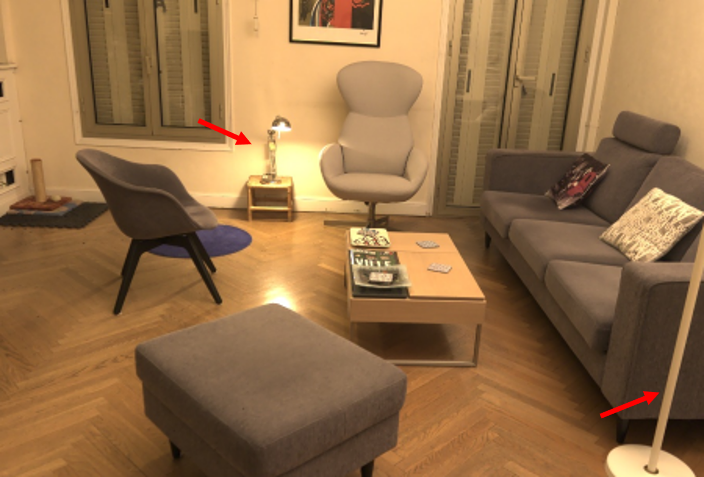} & \includegraphics[width=2.9cm]{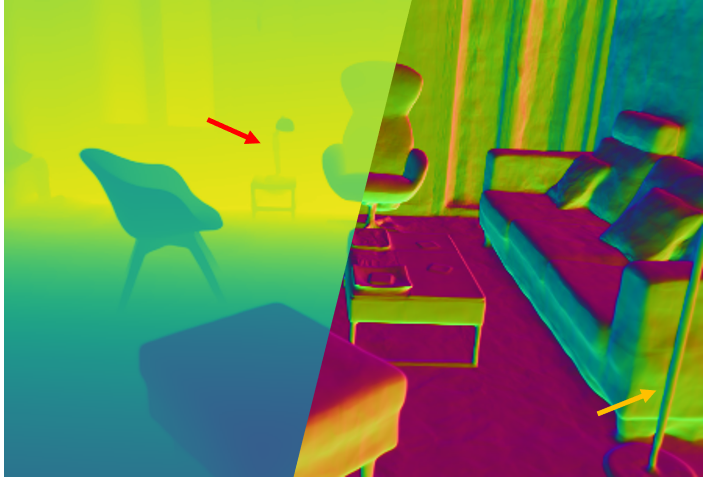} & \includegraphics[width=2.9cm]{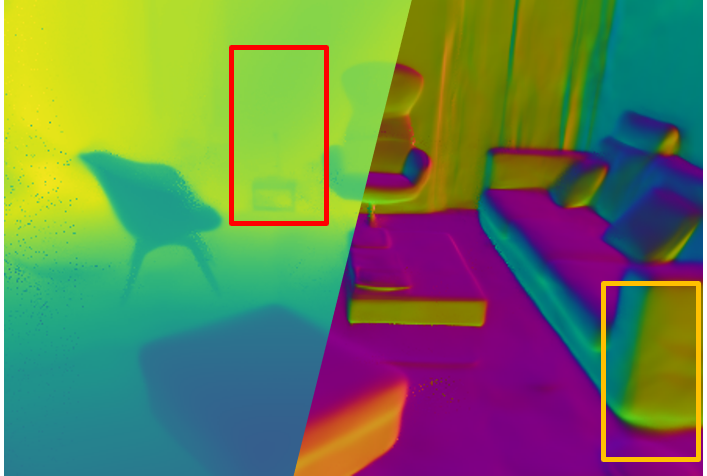} & \includegraphics[width=2.9cm]{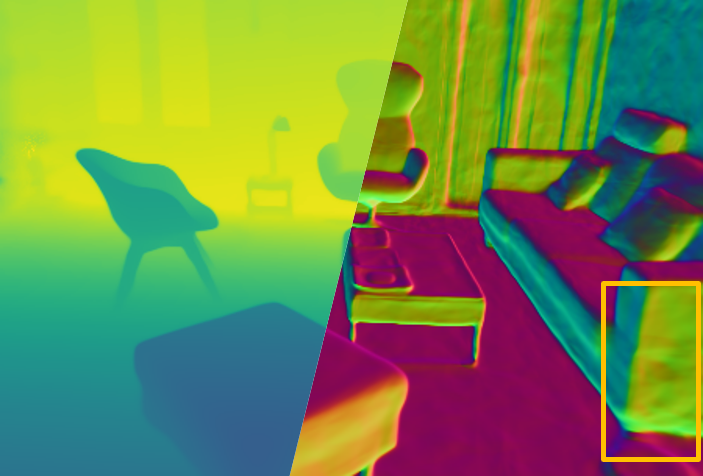} & \includegraphics[width=2.9cm]{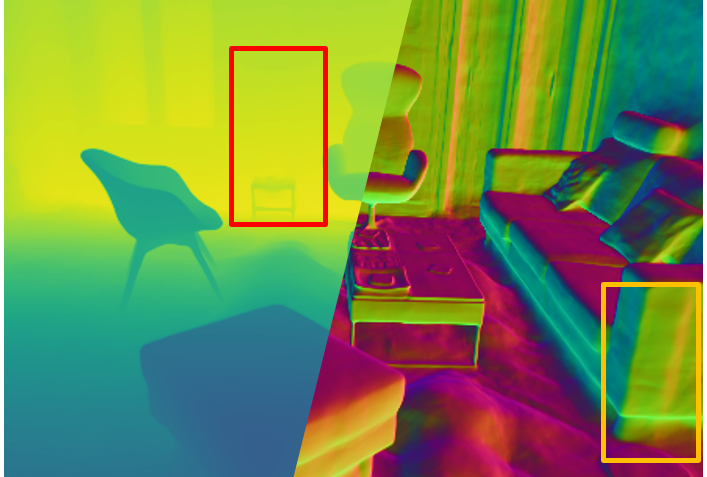} \\
    \end{tabular}
    \vspace{-1ex}
    \caption{\textbf{Qualitative comparisons of reconstructed depth map and normal map.}}
    \label{fig:sup_geo}
    \vspace{-1ex}
\end{figure*}
}

{
\setlength\tabcolsep{1.5pt}
\begin{figure*}[ht!]
    \centering
    \begin{tabular}{ccccccc}
         & GT & Ours & NeRF & Instant-NGP & NeuRIS & MonoSDF \\
         \multirow{2}{*}{\rotatebox{90}{Synthetic}} & \includegraphics[width=2.5cm]{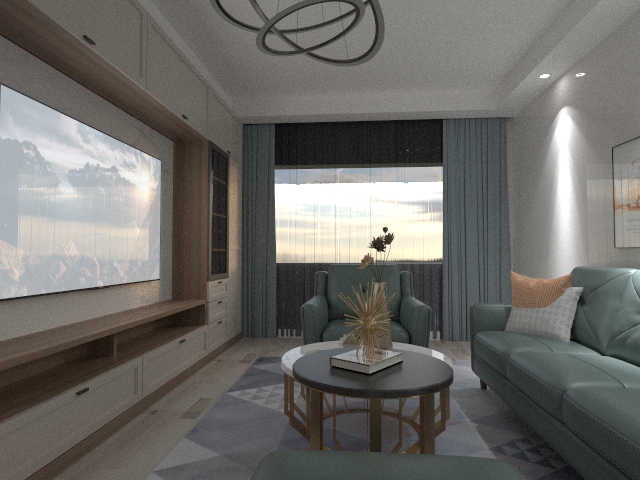} & \includegraphics[width=2.5cm]{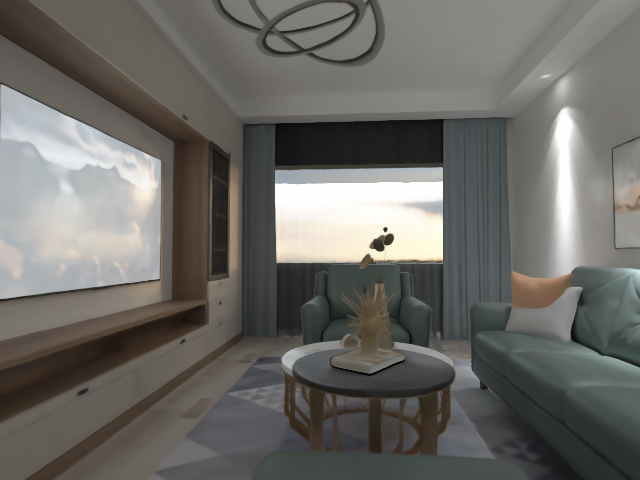} & \includegraphics[width=2.5cm]{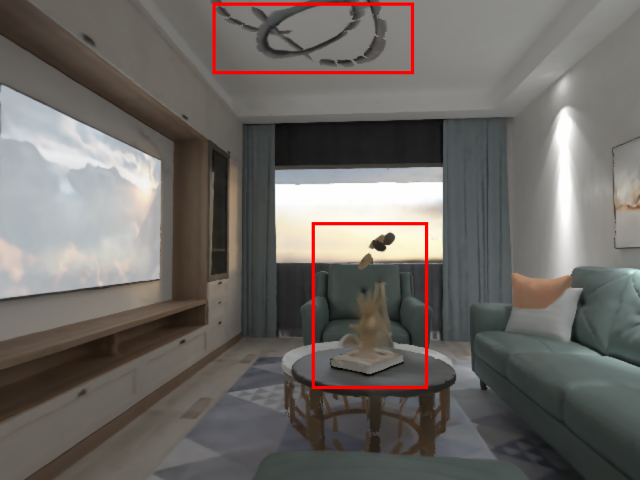} & \includegraphics[width=2.5cm]{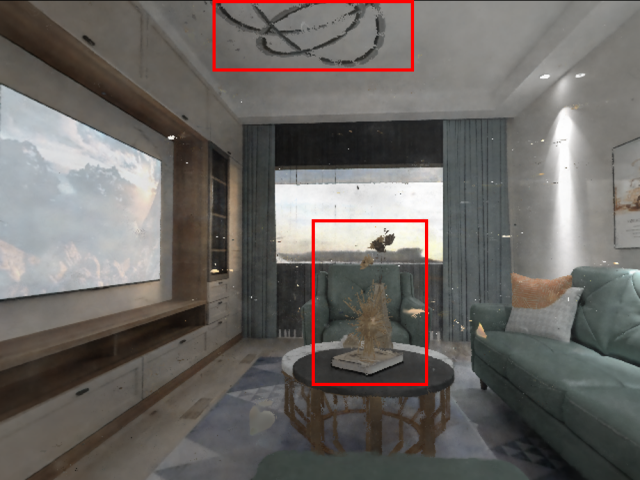} & \includegraphics[width=2.5cm]{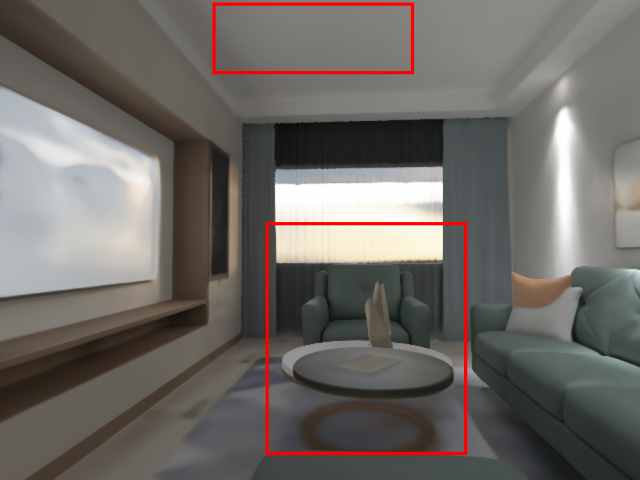} & \includegraphics[width=2.5cm]{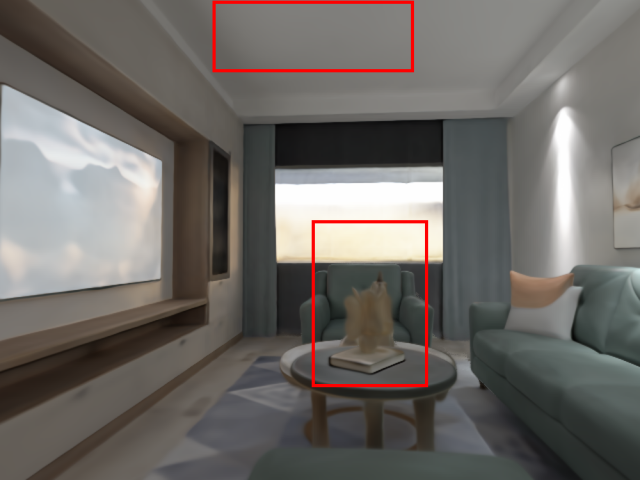}\\
         & \includegraphics[width=2.5cm]{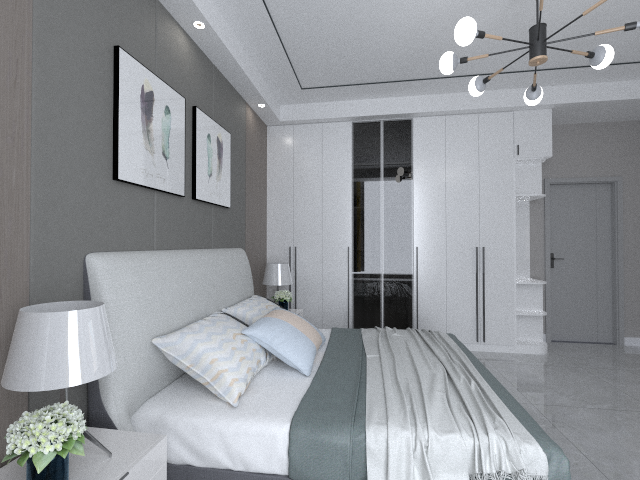} & \includegraphics[width=2.5cm]{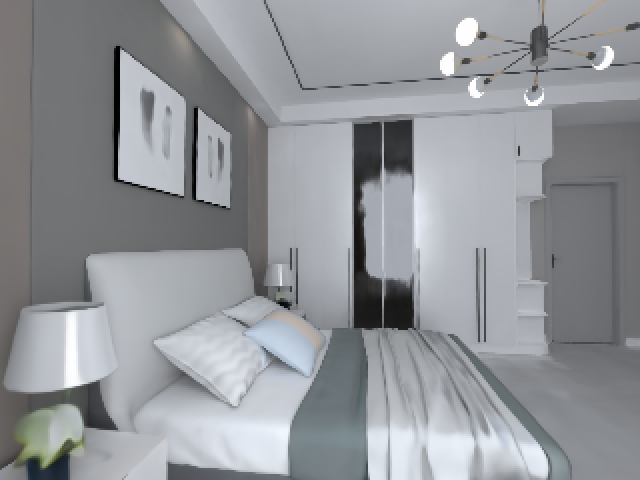} & \includegraphics[width=2.5cm]{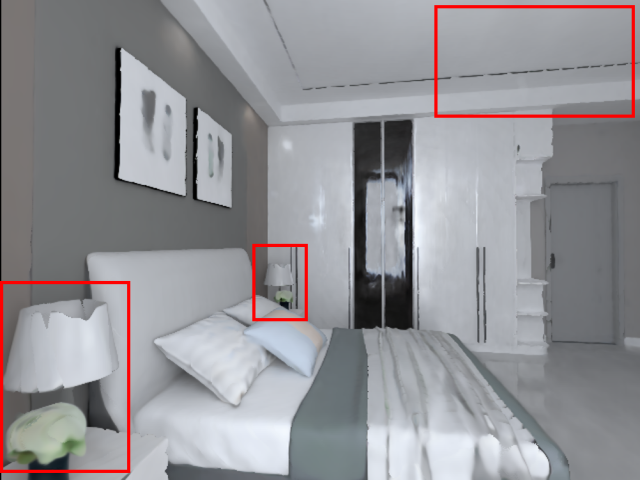} & \includegraphics[width=2.5cm]{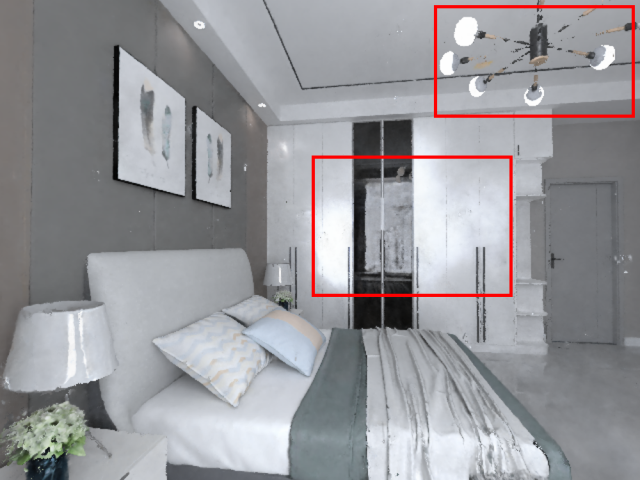} & \includegraphics[width=2.5cm]{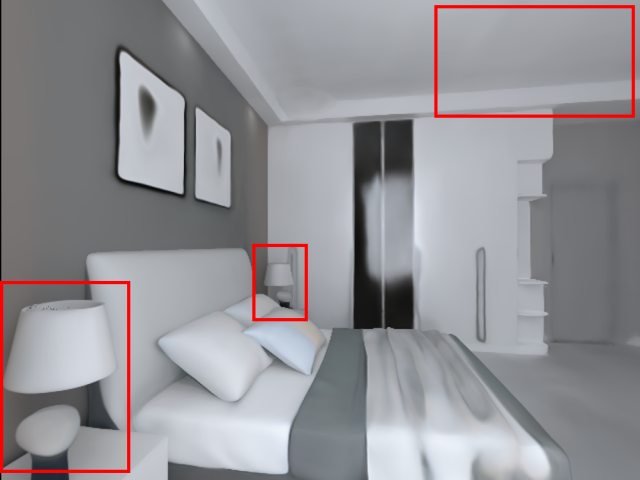} & \includegraphics[width=2.5cm]{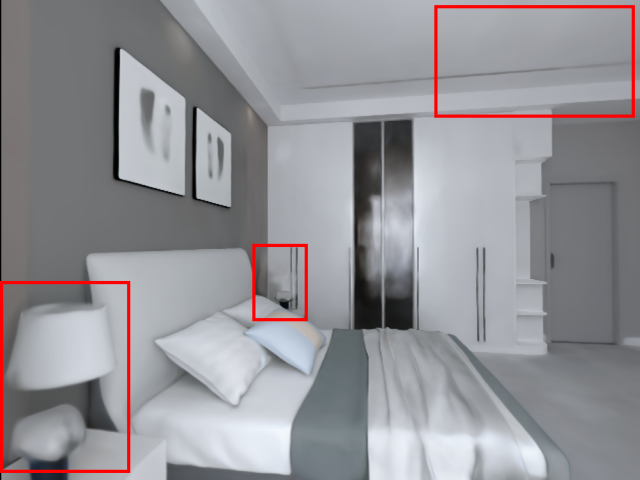} \\

         \multirow{2}{*}{\rotatebox{90}{Real}} & \includegraphics[width=2.5cm]{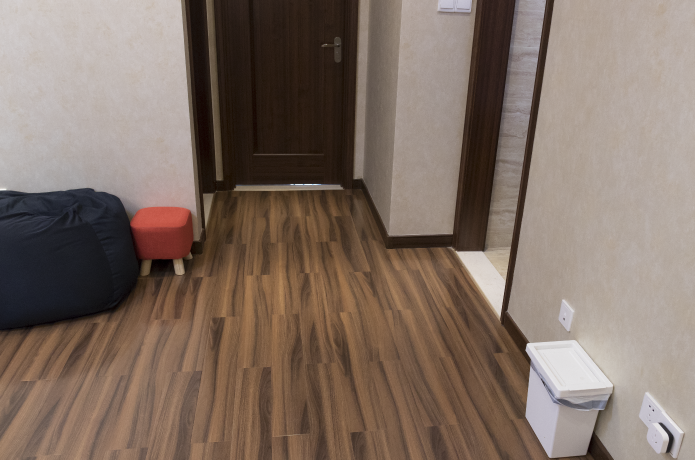} & \includegraphics[width=2.5cm]{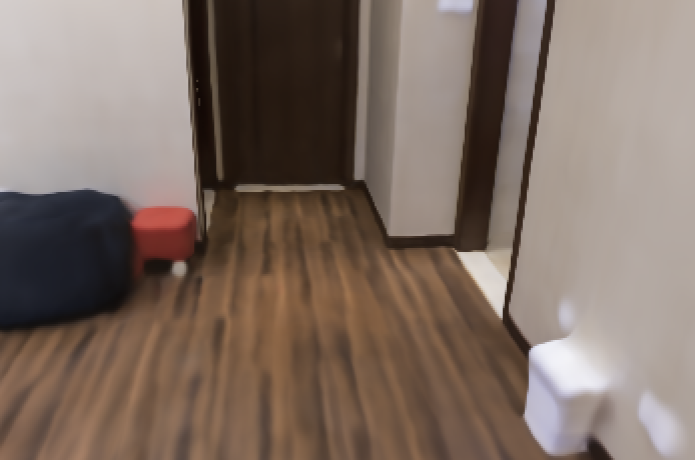} & \includegraphics[width=2.5cm]{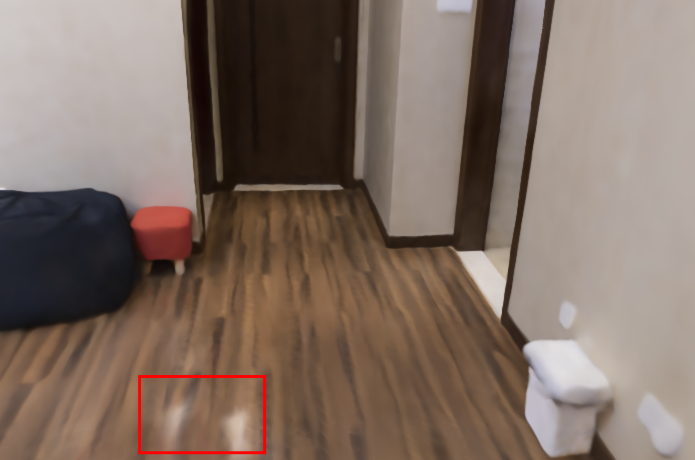} & \includegraphics[width=2.5cm]{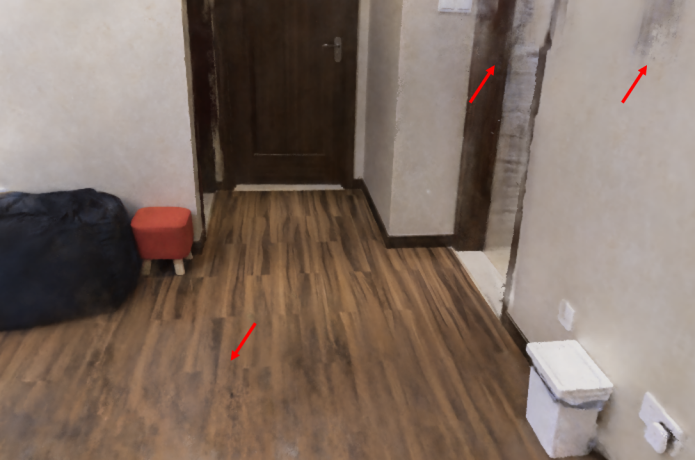} & \includegraphics[width=2.5cm]{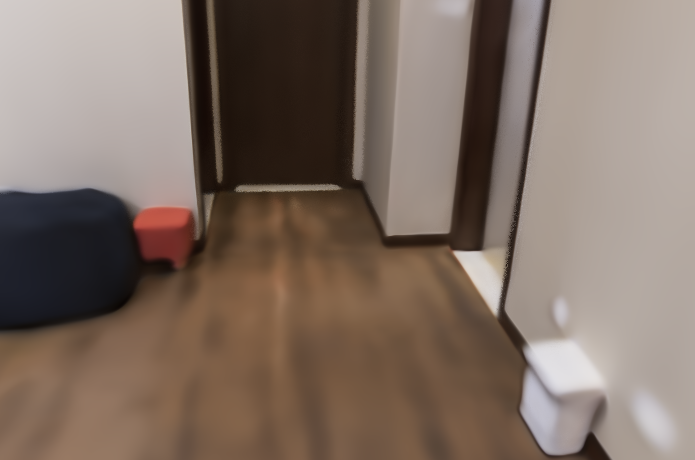} & \includegraphics[width=2.5cm]{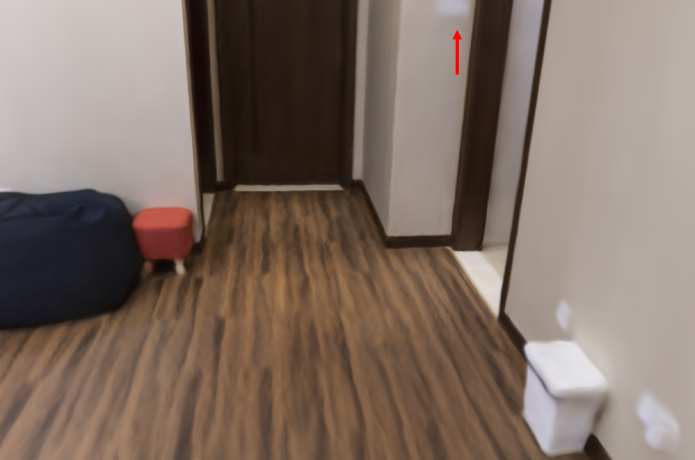}\\
         & \includegraphics[width=2.5cm]{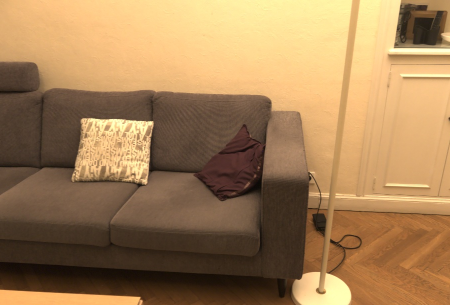} & \includegraphics[width=2.5cm]{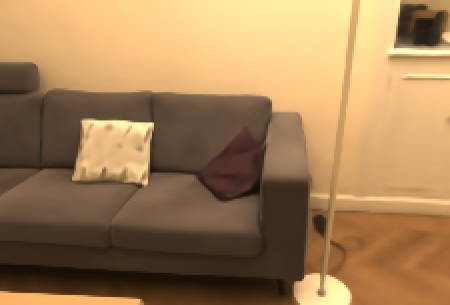} & \includegraphics[width=2.5cm]{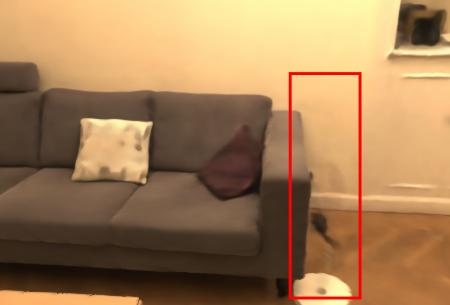} & \includegraphics[width=2.5cm]{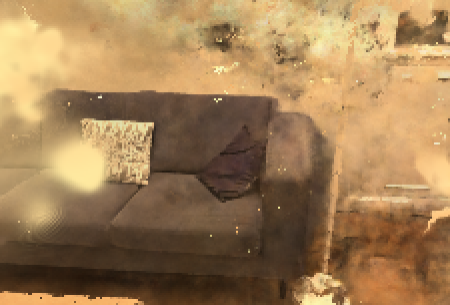} & \includegraphics[width=2.5cm]{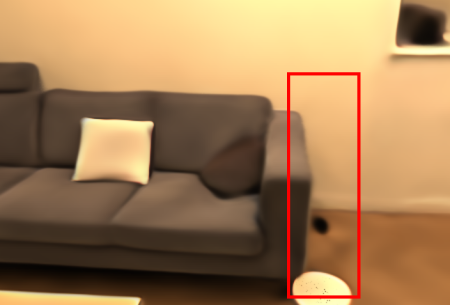} & \includegraphics[width=2.5cm]{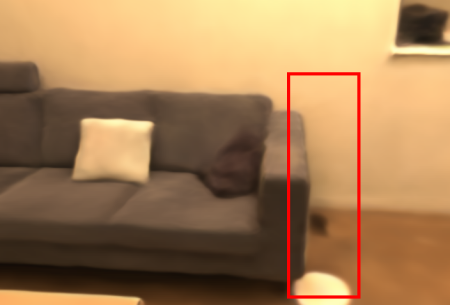} \\
    \end{tabular}
    \caption{\textbf{Qualitative comparisons of novel view synthesis} on synthetic data and real data. Zoom in for details.}
    \label{fig:sup_novel}
\end{figure*}
}

{
\setlength\tabcolsep{1pt}
\begin{table*}[ht!]
    \centering
    \caption{\textbf{Comparisons of per-scene novel view PSNR.}}
    \begin{tabular}{m{1.8cm}<{\centering}m{1.8cm}<{\centering}m{1.8cm}<{\centering}m{1.8cm}<{\centering}m{1.8cm}<{\centering}m{1.8cm}<{\centering}}\hline
         & Ours & NeRF & Instant-NGP & NeuRIS & MonoSDF \\\hline
        syn\_1 & 28.03 & 26.24 & 25.8 & 25.53 & 27.37 \\
        syn\_2 & 30.09 & 29.27 & 27.34 & 24.30 & 24.83 \\
        syn\_3 & 27.46 & 25.73 & 26.58 & 24.30 & 24.48 \\
        syn\_4 & 29.64 & 27.99 & 25.95 & 26.93 & 26.67 \\
        syn\_5 & 27.71 & 26.93 & 16.82 & 24.38 & 26.03 \\
        syn\_6 & 28.55 & 27.27 & 16.27 & 25.21 & 26.49 \\
        syn\_7 & 28.04 & 27.65 & 24.67 & 25.18 & 24.67 \\
        syn\_8 & 27.83 & 25.31 & 27.72 & 24.36 & 25.39 \\
        \textbf{mean} & \textbf{29.70} & 27.09 & 23.89 & 25.02 & 25.74 \\\hline
        real\_1 & 26.63 & 26.48 & 19.10 & 25.91 & 26.21 \\
        real\_2 & 28.01 & 27.21 & 26.32 & 23.87 & 24.38 \\
        real\_3 & 24.58 & 24.11 & 23.29 & 23.31 & 22.72 \\
        real\_4 & 21.39 & 20.86 & 19.23 & 19.82 & 20.05 \\
        \textbf{mean} & \textbf{25.15} & 24.66 & 21.99 & 23.22 & 23.34 \\\hline
    \end{tabular}
    \label{tab:sup_novel}
\end{table*}
}

\begin{table*}[ht!]
    \centering
    \caption{\textbf{Comparisons of per-scene normal angular error and depth $L_1$ loss.}}
    \begin{tabular}{cccccccc}\hline
         & \multicolumn{3}{c}{Normal-Angular-$L_1 \downarrow$} & & \multicolumn{3}{c}{Depth-$L_1 \downarrow$} \\\cline{2-4}\cline{6-8}
         & Ours & NeuRIS & MonoSDF & & Ours & NeuRIS & MonoSDF \\\hline
        syn\_1 & 0.040 & 0.051 & 0.036 & & 0.014 & 0.240 & 0.010 \\
        syn\_2 & 0.030 & 0.041 & 0.035 & & 0.019 & 0.331 & 0.048 \\
        syn\_3 & 0.054 & 0.080 & 0.073 & & 0.021 & 0.319 & 0.061 \\
        syn\_4 & 0.053 & 0.071 & 0.065 & & 0.068 & 0.312 & 0.103 \\
        syn\_5 & 0.064 & 0.096 & 0.058 & & 0.025 & 0.227 & 0.034 \\
        syn\_6 & 0.057 & 0.082 & 0.054 & & 0.051 & 0.355 & 0.043 \\
        syn\_7 & 0.057 & 0.071 & 0.076 & & 0.065 & 0.335 & 0.099 \\
        syn\_8 & 0.072 & 0.071 & 0.064 & & 0.016 & 0.274 & 0.033 \\
        \textbf{mean} & \textbf{0.053} & 0.070 & 0.058 & & \textbf{0.035} & 0.299 & 0.054 \\\hline
    \end{tabular}
    \label{tab:normal}
\end{table*}

{
\setlength\tabcolsep{1.5pt}
\begin{figure*}[ht!]
    \centering
    \begin{tabular}{cccc}
        Image & Ours & MonoSDF & NeuRIS \\
        \includegraphics[width=4cm]{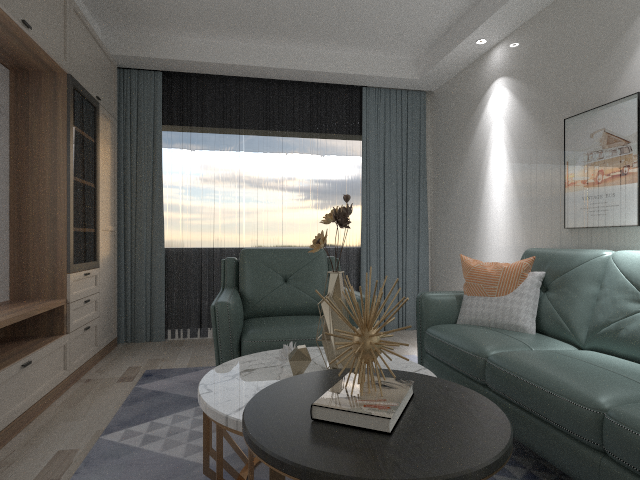} & \includegraphics[width=4cm]{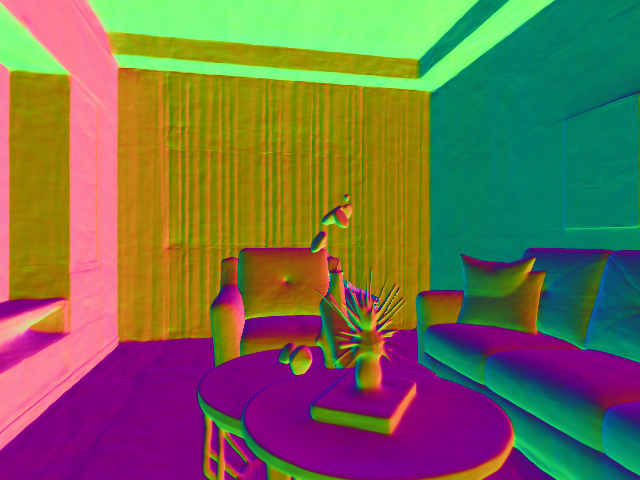} & \includegraphics[width=4cm]{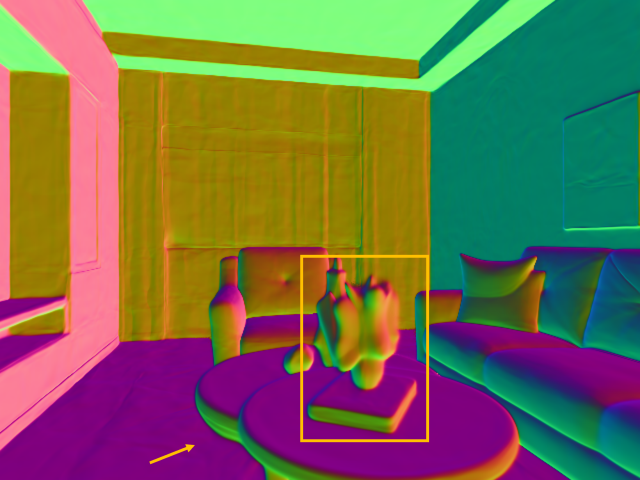} & \includegraphics[width=4cm]{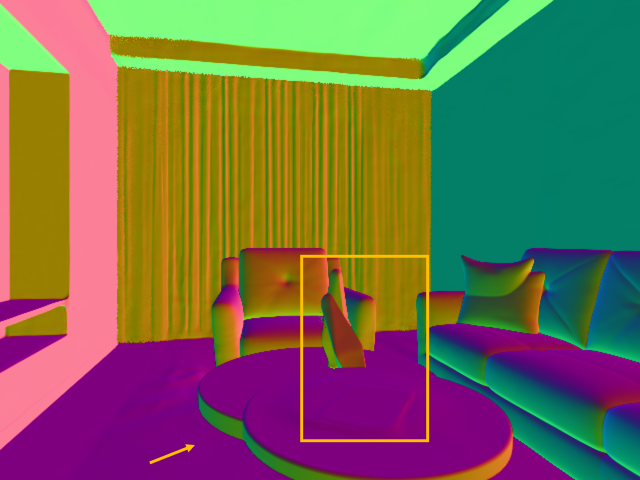} \\
        \includegraphics[width=4cm]{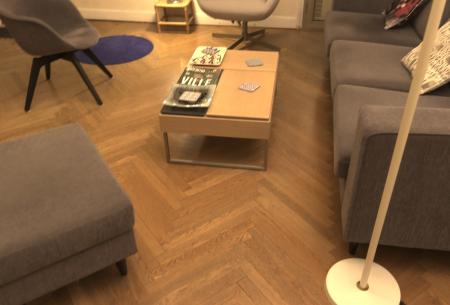} & \includegraphics[width=4cm]{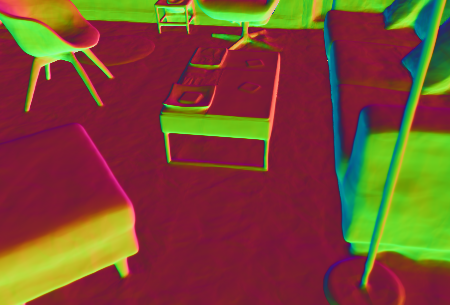} & \includegraphics[width=4cm]{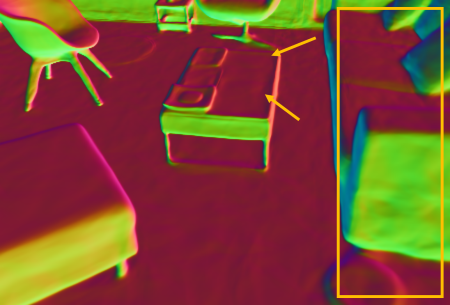} & \includegraphics[width=4cm]{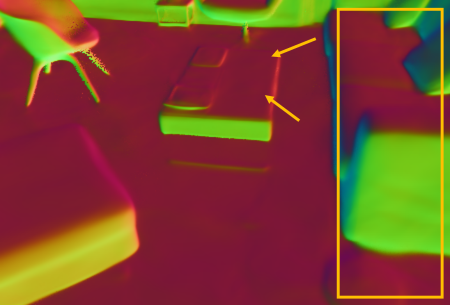} 
    \end{tabular}
    \caption{\textbf{Qualitative comparisons of normal estimation} on synthetic data and real data.}
    \label{fig:sup_normal}
\end{figure*}

        

\begin{figure*}[h]
    \centering
    \begin{tabular}{cccc}
        Image & $K_d$ & $K_s$ & $\rho$ \\
        \includegraphics[width=4cm]{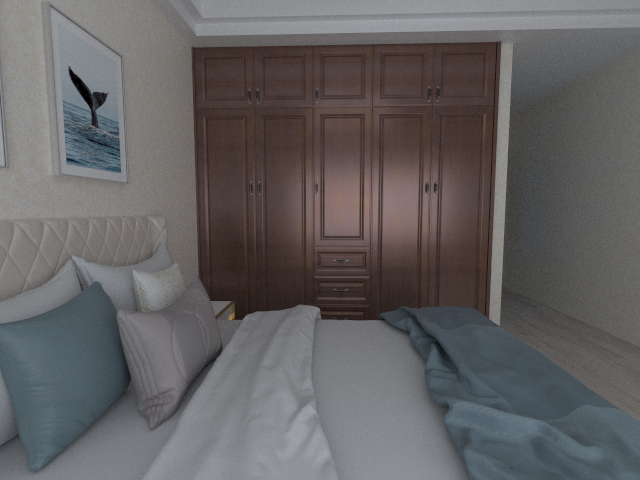} & \includegraphics[width=4cm]{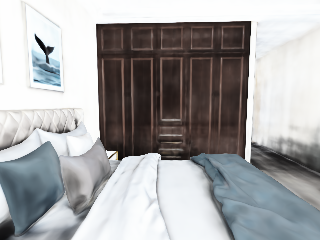} & \includegraphics[width=4cm]{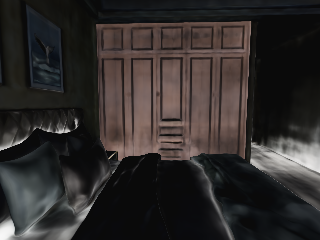} & \includegraphics[width=4cm]{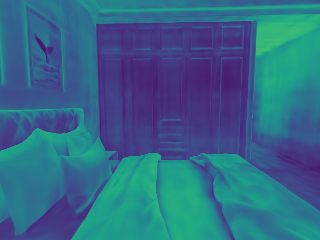} \\
        \includegraphics[width=4cm]{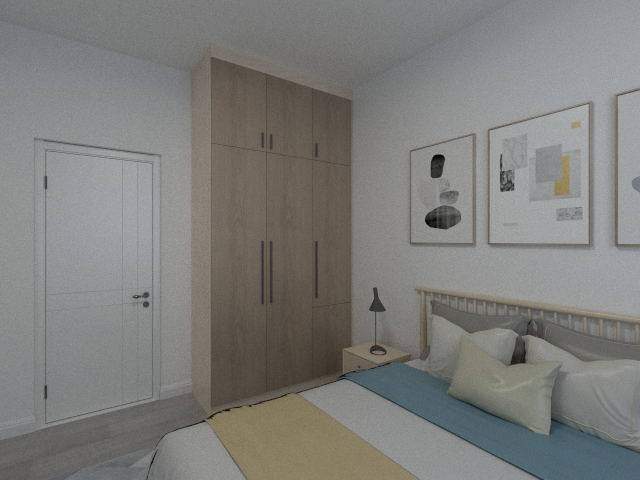} & \includegraphics[width=4cm]{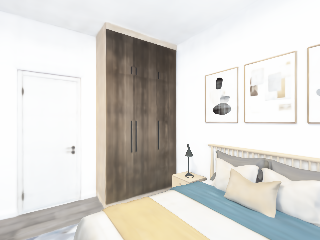} & \includegraphics[width=4cm]{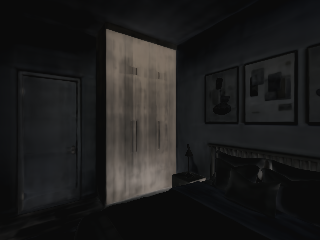} & \includegraphics[width=4cm]{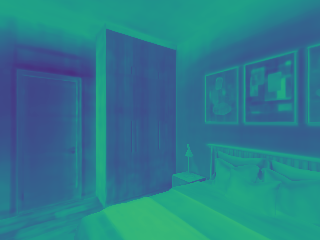} \\
    \end{tabular}
    \caption{\textbf{Qualitative results of decomposed materials.}}
    \label{fig:sup_mat}
\end{figure*}

\begin{figure*}[h]
    \centering
    \begin{tabular}{cccc}
        Orig. & Edited & Orig. & Edited \\
        \includegraphics[width=4cm]{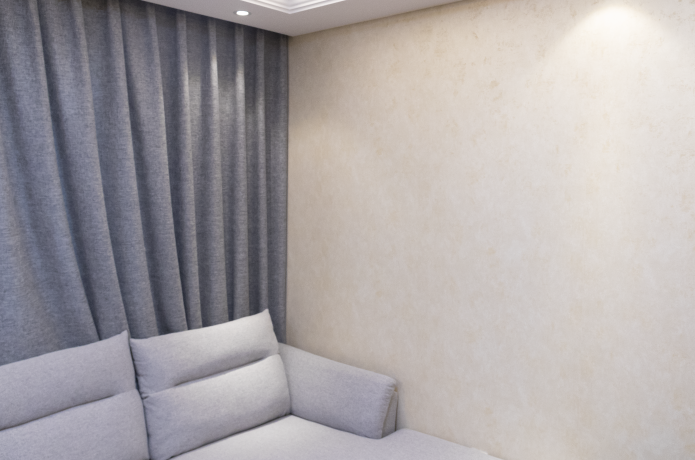} & \includegraphics[width=4cm]{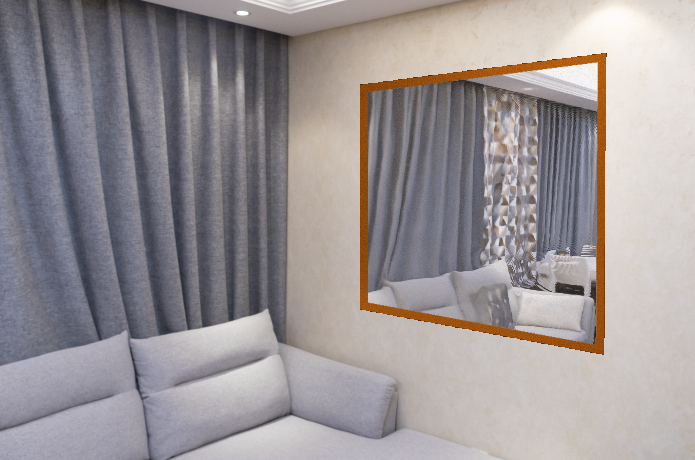} & \includegraphics[width=4cm]{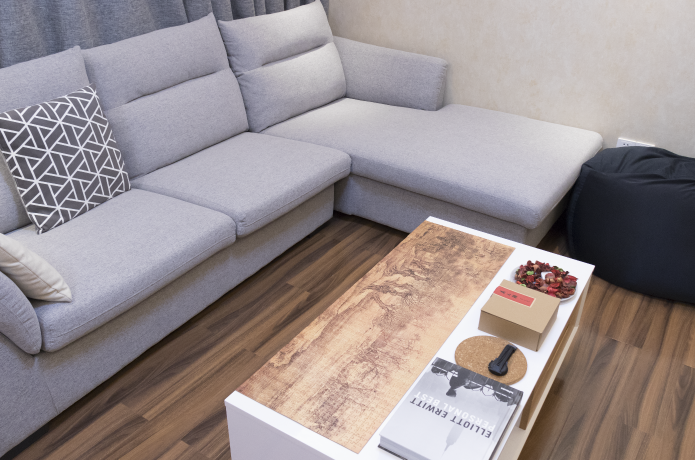} & \includegraphics[width=4cm]{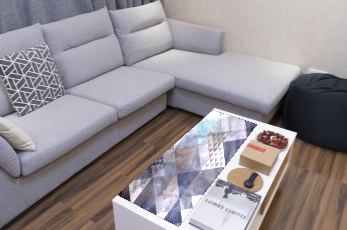} \\
        \includegraphics[width=4cm]{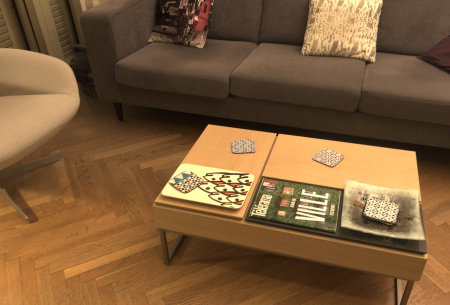} & \includegraphics[width=4cm]{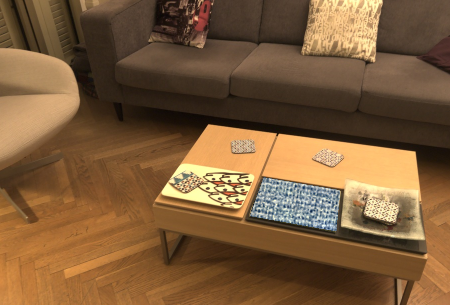} & \includegraphics[width=4cm]{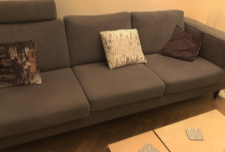} & \includegraphics[width=4cm]{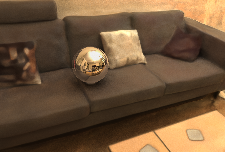} \\
        \includegraphics[width=4cm]{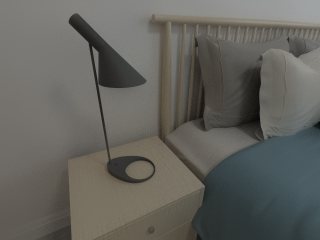} & \includegraphics[width=4cm]{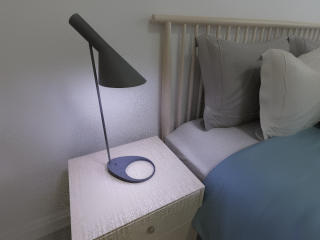} & \includegraphics[width=4cm]{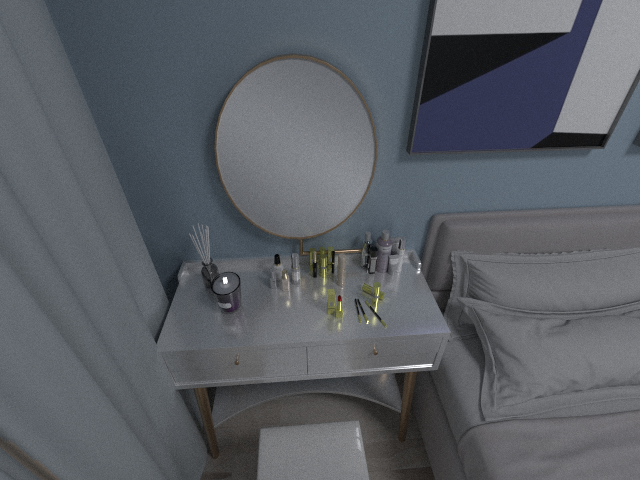} & \includegraphics[width=4cm]{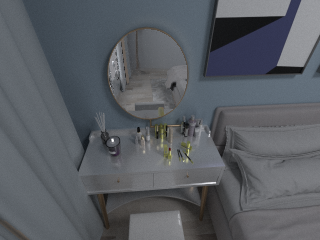}
    \end{tabular}
    \caption{\textbf{Qualitative results of scene editing and relighting.}}
    \label{fig:sup_edit}
\end{figure*}

}

\section{Additional Experimental Results}\label{sec:result}

\paragraph{Novel view synthesis.} \cref{tab:sup_novel} displays quantitative results of per-scene novel view PSNR. It turns out that our method outperforms all of the baselines, benefiting from our precise reconstruction of small objects and proper handling of shape-radiance ambiguity. Qualitative results are presented in \cref{fig:sup_novel}. NeRF, NeuRIS and MonoSDF fails to reconstruct small objects such as chandeliers and lamp poles, while NeRF and Instant-NGP also sufffers from fractured reconstruction results. While Instant-NGP usually capture most high-frequency details, it likely overfits to single-view radiance and fails to ensure multi-view geometry consistency in indoor scenes, leading to poor novel view synthesis results with floating artifacts.

\paragraph{Geometry Reconstruction.} \cref{tab:normal} displays quantitative comparisons of per-scene normal angular $L_1$ error and depth $L_1$ error between our method and baselines. The definition of normal angular $L_1$ error is
\begin{equation}
    \mathcal{L}_\mathrm{normal} = \| 1 - \hat{N}\cdot N \|_1
\end{equation}
Our method also outperforms NeuRIS and MonoSDF, indicating superior 3D reconstruction quality. \cref{fig:sup_normal} presents qualitative comparisons of the reconstructed normal maps. Our method can even recover high-frequency details on geometry, such as the spikes on the ball.

\paragraph{Material Decomposition.} 
\cref{fig:mat,fig:sup_mat} presents qualitative results of the decomposed diffuse albedo $K_d$, specular albedo $K_s$ and roughness $\rho$. The material parameters are not directly supervised by GT labels, but we produce plausible results.

\paragraph{Scene editing.} With the intrinsic decomposition results, we can enable photo-realistic scene editing tasks such as material editing and relighting. \cref{fig:sup_edit} shows qualitative results of scene editing results in both real and synthetic data. We explore mirror insertion (top-left and bottom-right), texture editing (top-right and mid-left), object insertion (mid-right) and relighting (bottom-left). Note that the edited specular reflections (on mirrors and inserted metal ball) are consistent with the surroundings. On account of our raytracing algorithm, our method is capable of casting shadows of the inserted object (see the shadows of the inserted ball on the sofa).